\documentclass[11pt]{article} 

\newcommand{\anon}[2] {#2}

\usepackage[utf8]{inputenc}
\usepackage{caption, subcaption}
\usepackage{subcaption}
\usepackage{booktabs}
\usepackage{graphicx}
\usepackage{listings}
\usepackage{enumitem}
\usepackage{xcolor}
\usepackage{amsmath}
\usepackage{amsfonts}
\usepackage{amssymb}
\usepackage{pifont}
\usepackage{url}
\usepackage{hyperref}
\usepackage{multirow}
\usepackage{cite}

\def\nrev#1{#1}

\def\mrrs{MRR}
\newcommand{\pnleg} {${\rm PN}^{22}$}
\newcommand{\pnnew} {${\rm PN}^{23}$}
\newcommand{\pnswe} {${\rm PN}_{}^{23\text{F}}$}

\newcommand{\mycomment}[1]{}

\begin{document}

\title{Floralens: a Deep Learning Model \\ for the Portuguese Native Flora}

\author{António Filgueiras$^{1}$,  Eduardo R. B. Marques$^{1,2,*}$, \\ Luís M. B. Lopes$^{1,2}$, Miguel Marques$^{1}$, Hugo Silva$^{1}$
\\ \\
$^{1}$Department of Computer Science \\ Faculty of Sciences, University of Porto \\
$^{2}$CRACS/INESC-TEC \\
$^{*}$Corresponding author: ebmarques@fc.up.pt
}

\date{}

\maketitle

Machine learning has become a crucial tool for image-based species identification in citizen science initiatives. However, high-accuracy models  specific to particular regions or countries are rare. This results from the lack of high-quality datasets for these geographies, combined with inadequate documentation of current citizen science projects, which together hinder reproducibility.  
%
We present a high-quality dataset of Portuguese native flora and a deep learning model named Floralens trained on this dataset. We focus on using accessible, off-the-shelf deep learning tools and a fully transparent and reproducible methodology.
%
We compiled a dataset based on research-grade images from the Sociedade Portuguesa de Botânica and supplemented it with additional data from the Global Biodiversity Information Facility (GBIF). Using Google’s AutoML Vision service, we trained and evaluated a CNN-based classification model.
%
The resulting model achieves an accuracy superior to some of the state-of-the-art CNN-based models from Pl@ntNet, a leading citizen science project, and is integrated into the public website of project Biolens, which hosts local models for multiple taxa in Portugal.
%
Our findings demonstrate that carefully curated datasets combined with cloud-based machine learning tools can yield lightweight models for identifying regional flora with accuracies comparable to those of state-of-the-art platforms. The complete dataset and Python notebooks used in this study are publicly available.


\anon{}{
\section{Introduction}
\label{sec:introduction}

Advances in processing speed, storage capacity, and imaging sensors for mobile devices have paved the way for citizen science applications~\cite{citizen_science}.
Prominent examples are initiatives for nature observation that rely on the photographic recording of animals, plants, and fungi in their natural habitats. Several of these applications employ deep learning models for automated species identification from user-contributed images~\cite{inaturalist-app,observation-org-model,waldchen2018,floraincognita-app,plantnet-model}. The data gathered by such projects are valuable to scientists, from professional taxonomists to ecologists studying the effects of human activities on biodiversity~\cite{tuia2022,christin2019}. 

In the Biolens project~\cite{biolens}, we aim to develop lightweight, locally deployable identification models for web-based and mobile applications capable of operating offline and supporting delayed data sharing—features that are largely absent from most existing citizen science platforms.  Moreover, detailed descriptions of their methodological approaches are scarce in the literature (cf. Section~\ref{sec:related}), limiting the reuse and reproducibility of accumulated knowledge in new projects. We therefore developed a streamlined methodology to construct datasets for native Portuguese species and to train accurate convolutional neural network (CNN)–based models using off-the-shelf machine learning tools.
Currently, we have developed four models: Lepilens and Mothlens (for butterflies and moths, together covering the order \emph{Lepidoptera}); Dragonlens (for dragonflies and damselflies, covering the order \emph{Odonata}); and the most recent addition, Floralens (targeting the \emph{Plantae} kingdom).

\begin{figure*}[t]
\centering    \includegraphics[width=0.9\linewidth,keepaspectratio=true]{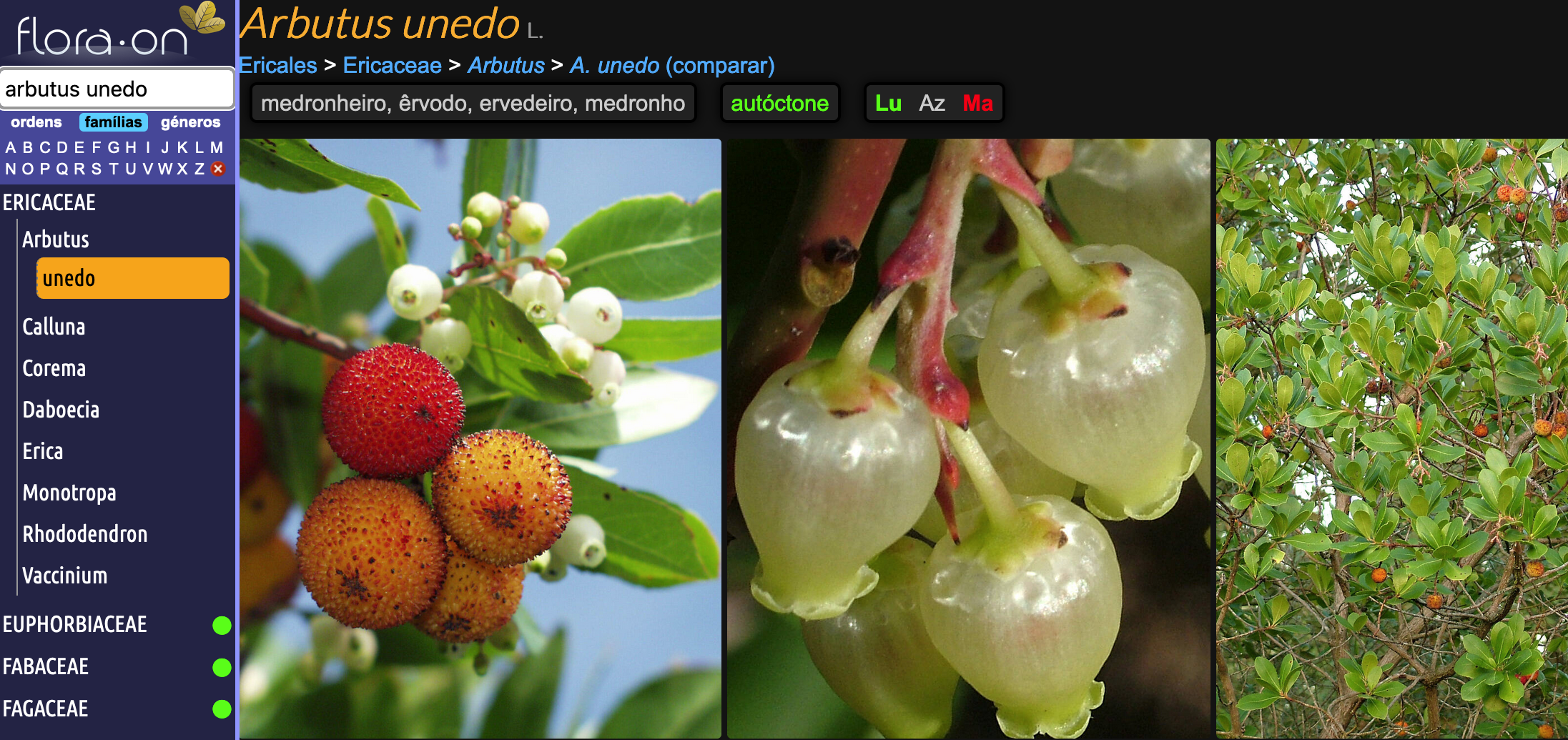}
    \caption{Detail of the FloraOn web application.}
    \label{fig:floraon}
\end{figure*}

This paper describes the derivation of Floralens, a high-accuracy machine learning model for automated taxonomic identification of the Portuguese native flora. The work is anchored in the FloraOn dataset provided by the Sociedade Portuguesa de Botânica~\cite{flora-on}, available online via a web application (Figure~\ref{fig:floraon}) and as a contributed dataset in the Global Biodiversity Information Facility (GBIF)~\cite{gbif-paper}. Although this dataset contains a relatively small number of images per species, each image is of high quality and has been identified by experienced taxonomists. This list served as the reference for building the dataset.

The methodology used to derive the Floralens model has two key characteristics: (a) the use of data from research-grade public repositories for dataset construction, and (b) the use of Google’s AutoML Vision (GAMLV) to train the model. This methodology is briefly described in a short scientific outreach article~\cite{cienciaelementar22}, in MSc theses~\cite{marques2021,filgueiras2022} (covering Floralens), and a BSc project report~\cite{mamede2020} (covering Lepilens). Our analysis shows that the Floralens model outperforms CNN-based models from state-of-the-art citizen science projects such as Pl@ntNet. In addition to these baseline results, we examine the effects of using multiple images per specimen and of species' geographical distribution on model accuracy.

The main contributions of this paper are as follows:
\begin{itemize}
\item a dataset for the Portuguese flora available from Zenodo~\cite{dataset};
\item a high-accuracy CNN-based model for the Portuguese flora publicly available via web and mobile applications; 
\item a thorough quantitative evaluation of the model and comparison with those from Pl@ntNet;
\item and a complete set of Python notebooks used to perform the analysis presented in this paper, also available from Zenodo.
\end{itemize}

The remainder of this paper is structured as follows. Section~\ref{sec:related} discusses the current state-of-the-art regarding automatic taxonomic identification based on deep learning. Section~\ref{sec:datasets} describes the construction of the dataset used in this study, the model's generation using GAMLV, and the evaluation metrics used to gauge its accuracy. Section~\ref{sec:results} presents the results obtained with the model and its refinements, such as using multiple images of the same specimen for classification and ecological indicators such as the geographical distribution of species. A detailed performance comparison with Pl@ntNet is also presented. Section~\ref{sec:artifacts} describes the software artifacts and datasets produced in the scope of this work. Finally, Section~\ref{sec:conclusions} discusses the main findings of this study and proposes future research directions.

\section{State-of-the-Art}
\label{sec:related}

\nrev{The advent of deep learning enabled the development of the first automated tools for identifying plant species from input images~\cite{lee2015, heredia2017, sun2017}. Subsequent refinements quickly reached a point where automated identification rivaled those made by specialists~\cite{bonnet2018}, highlighting AI's transformative role~\cite{nia}. Models based on deep learning are currently central tools in major citizen science platforms such as iNaturalist~\cite{inaturalist-app}, Observation.org~\cite{observation-org-model}, and Pl@ntNet~\cite{plantnet-model}, and are integrated into web and mobile applications. Most of these models are derived using convolutional neural networks~\cite{cnn1,cnn2} and, recently, vision transformers (ViTs)~\cite{transformers,transformers2,plantnet_api_announcement}. Citizen science platforms generate relevant resources such as curated datasets, which in turn are often made available to the public through biodiversity data portals, notably GBIF~\cite{gbif-paper}. These datasets enable the development of other ML models, as is the case of Floralens, which, in addition to FloraOn, uses data sampled from GBIF datasets provided by iNaturalist, Observation.org, and Pl@ntNet (cf. Section~\ref{sec:datasets}).} However, the dataset construction and model derivation methodologies used by these citizen science platforms are seldom documented in detail. For instance, \cite{inaturalist-app} discusses machine learning only from an end-user perspective, while additional information available online~\cite{inat-cv} is fragmentary. A brief overview of Pl@ntNet’s model derivation process appears in~\cite{plantnet-model} and is presented in a slightly more detailed form in~\cite{plantnet-model-2}. Likewise, \cite{observation-org-model} offers only a short abstract describing the ObsIdentify~\cite{obsidentify} mobile application, and both the app and the site interface are enabled by an API~\cite{nia,nia2} whose underlying model remains undocumented. Moreover, extended access to the classification services in these platforms for research is limited because, in some cases, the service is not exposed by the API or, commonly, requires a paid subscription for high-volume requests.

Developing CNN-based or other ML models typically requires expert knowledge, complex configuration, and specialized code, along with substantial computational resources for data storage and model training. The strain is more acute when dealing with large datasets, as in the case of Floralens, which uses approximately 300,000 images to derive a model. AutoML cloud services,
such as GAMLV, which we used in this work, address these challenges by automating several steps of model creation. Similar machine-learning-as-a-service (MLaaS) platforms include Amazon Rekognition~\cite{rekognition}, Apple Create ML~\cite{createml}, and Azure AutoML~\cite{azure_automl}. These services provide efficient infrastructure with specialized hardware (e.g., GPUs or TPUs), enabling rapid model training. Models derived using such platforms are increasingly referenced in various fields—including biology~\cite{choudhary2023non}, medicine~\cite{korot2021code,borkowski2019google}, agriculture~\cite{MALOUNAS2024100437}, and engineering~\cite{liang2023integrating}. Comparative studies~\cite{choudhary2023non,korot2021code,borkowski2019google} generally confirm the strong performance of GAMLV-derived models. In this work, we adopted GAMLV because we had free access to the service and computational resources through Google Cloud’s Research Credits program.

This is a rapidly evolving research area driven by developments in machine learning and cloud computing infrastructure. A recent trend involves using metadata associated with observations to improve the accuracy of the models. For instance, Pl@ntNet developed regional models by dividing the globe into large biogeographic domains~\cite{ai-geospecies} based on regional floras such as WCVP/Kew~\cite{WCVP}, and incorporated metadata on plant parts depicted in images (e.g., flowers, leaves, stems). Rzanny et al.~\cite{rzanny2019} adopted a similar strategy, using five standardized image perspectives per specimen, training separate models for each, and combining their outputs to improve classification accuracy. M{\"a}der et al.~\cite{floraincognita-app} (Flora-Incognita) employed CNNs to classify images of German native plants, complemented by a deep feedforward network predicting species occurrence by location and season. The outputs of both models are fused via a recurrent neural network. More recently, Brun et al.~\cite{florid} (FlorID) used data-efficient image transformers trained on images annotated with metadata such as location, time, and plant part. Like Flora-Incognita, FlorID integrates ecological predictors (e.g., elevation, season) with image-based models to refine predictions. Persistent challenges remain, including the limited accuracy of these models in identifying endangered species~\cite{mindyourapp}.
Floralens is closer in philosophy to regional flora projects such as Flora-Incognita~\cite{floraincognita-app} and FlorID~\cite{florid}.
Another trend that has generated considerable interest is the development of global models trained on large-scale datasets such as iNat Challenge~\cite{van2018} and PlantCLEF/LifeCLEF~\cite{goeau2017,goeau2022}, consisting of millions of images representing tens of thousands of species among the roughly 300,000 known worldwide.

\section{Methods}
\label{sec:datasets}

In this section, we describe the creation of the Floralens dataset, the derivation of the Floralens model, and the evaluation metrics used to assess the model’s performance.

\subsection{Data}
Our universe of species was the FloraOn catalog as of November 2021, containing 2,712 species of native Portuguese flora. We began defining the dataset by gathering a large collection of images for each species in our list. The images were extracted from curated datasets available from GBIF. We sampled the collection to include between 50 and 200 images per species. This ensured that each species was sufficiently represented while avoiding over-representation, which could introduce bias in the resulting models. These bounds were defined based on our prior experience building datasets for other biological taxa (cf. Section~\ref{sec:artifacts}).

The FloraOn repository contains geolocated records of flora species with associated images. The image data are relatively broad in scope, covering 78\% of the entire catalog (2,127 out of 2,712 species). However, it has a limited volume since, on average, there are only 11 images per species, meaning that our 50-image lower threshold is not met for any species.  Hence, to adequately populate the Floralens dataset, we retrieved the FloraOn images and also included image data from three publicly available datasets hosted on GBIF by citizen science platforms: iNaturalist~\cite{inat-gbif}, Observation.org~\cite{ObsOrg-gbif}, and Pl@ntNet~\cite{PlantNet-gbif}. This approach is also used by other platforms when building their datasets, e.g., FlorID~\cite{florid}.

The GBIF datasets provide validated observation data and associated images, originally submitted by users of the respective platforms. However, the validation process differs among data sources, as discussed below. For each dataset and each species in our universe, we used GBIF portal queries to obtain observation records in the Darwin Core Archive format~\cite{gbif-darwin}. Each record corresponds to an observation of a specimen by citizen scientists or experts. The observation is composed of a taxonomic identification, date, geographical location, and one or more images. \nrev{The GBIF portal queries covered continental Europe (including Great Britain, Ireland, and the Mediterranean islands). This broader scope was chosen because preliminary analysis showed that restricting queries to Portugal or the Iberian Peninsula would yield too little data. Fortunately, many species of the Portuguese flora are widely distributed.}

\begin{table}[t]
\centering
\caption{Raw data and derived dataset after sampling (\#I: image count; \#S: species count; $\ge 50$ I: species
with more than 50 images).}\label{t:dataset}
{ \small
\begin{tabular}{@{}lrrrr@{\quad\quad\quad}rrr@{}}
\toprule
& \multicolumn{4}{c}{Raw data} & \multicolumn{3}{c}{Dataset} \\
Source  &  \#I & \%I & \#S & $\ge 50$~I &  \#I & \%I & \#S  \\
\midrule
FloraOn         &                  22,869  & 0.5 &         2,127 &   0     &  15,191 & 5   &            1,397             \\
iNaturalist     &                2,753,167 & 66.2   &     2,066 &  1,431  &  90,127 &   31   &       1,358              \\
Observation.org &                 823,389 &  19.8  &    1,816 &  1,114     &  85,746 &  29   &        1,093              \\
Pl@ntNet        &                 515,950 &  12.4   &    1,495 &   735      &  102,537 &  35 &          1,373             \\
\midrule
Total   &                4,154,895 &      &     2,539 &  1,678           &  293,601 &   100   & 1,678     \\
\bottomrule
\end{tabular}
}
\end{table}

The raw data from all image sources are listed in Table~\ref{t:dataset} (left), along with the characterization of the Floralens dataset (right) that results from sampling the raw data. The corresponding histograms, depicting the number of species versus the number of images, are illustrated in Figure~\ref{fig:species-counts}. In the raw data, more than 4 million images were available for consideration, covering 2,539 species (93\% of the FloraOn catalog). Only 0.5\% of these images are from FloraOn, and approximately two-thirds are taken from iNaturalist. Moreover, only 1,678 species reached our lower bound threshold of 50 images (61\% of the FloraOn catalog). This scarcity for some species can be due to subjective issues, such as the visual attractiveness of the plant, e.g., having a showy flower, or it can be a real effect, reflecting its rare status in the wild. This effect gives the raw data a long-tailed distribution (Figure~\ref{fig:species-counts-raw}, shown in logarithmic scale).

The Floralens dataset was derived by sampling the raw data as follows. First, we filtered out species with fewer than 50 images. Then, for each of the remaining species, we sampled up to 200 images from the datasets, prioritizing data sources in the following order: (1) FloraOn; (2) Pl@ntNet; (3) Observation.org, and (4) iNaturalist. That is, for the 50-200 image target for each species, we use up as many images as possible from FloraOn first, then from Pl@ntNet, and so on. 

\begin{figure*}[t!]
\centering
    \begin{subfigure}{0.45\linewidth}
        \centering
        \includegraphics[width=\linewidth]{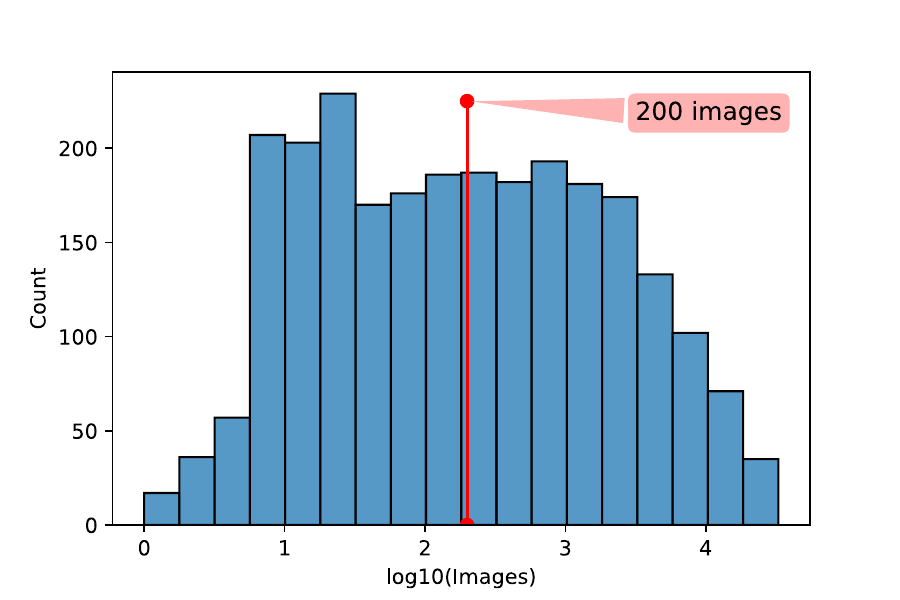}
        \subcaption{Raw data.}
        \label{fig:species-counts-raw}
    \end{subfigure}
    \begin{subfigure}{0.45\linewidth}
        \centering
        \includegraphics[width=\linewidth]{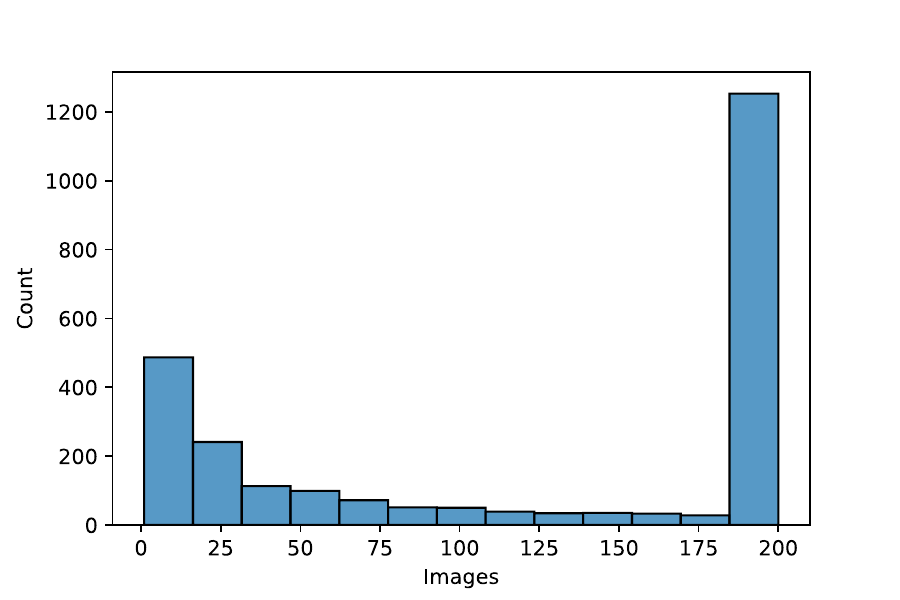}
        \subcaption{Floralens dataset.}
        \label{fig:species-counts-rebalanced-log}
    \end{subfigure}
    \caption{Dataset histograms ($x$-axis: number of images; $y$-axis: number of species).}
    \label{fig:species-counts}
\end{figure*}

This prioritization reduces identification errors by considering the curation rigor of each data source. FloraOn is curated by botanists and features high-quality images. These often feature subtle details that help secure the identification of a species. Pl@ntNet data undergoes a curation process that involves machine learning, contributors' reputation scores, and geographically-referenced species verification~\cite{PlantNet-gbif}.  Observation.org data can result from automatic validation through image recognition coupled with a check for other approved observations in the geographical vicinity, or through an expert volunteer when
automated validation fails~\cite{ObsOrg-gbif,ObsOrg-validation}. Finally, iNaturalist identifications result from a crowd-sourcing effort whereby a ``research-grade" identification for an image can be obtained from the consensus of just two citizen scientists~\cite{inat-gbif,inat-validation}.
We use images regardless of plant feature type, as this metadata was not available from the curated datasets. relying instead on the curation processes of the contributing platforms, which filter out inadequate submissions (e.g., out-of-focus or low-resolution images).

This procedure yielded the Floralens dataset, comprising approximately 300,000 labeled images across 1,678 species. As illustrated in Figure~\ref{fig:species-counts-rebalanced-log}, there are 200 images or very close to it for most of the species. The image count is 200 for 67\% (1,128) of the species, 150 or higher for 79\% (1,323), and 100 or higher for 86\% (1,449). The data source prioritization scheme led to a more significant fraction of FloraOn images in comparison to the raw data (the fraction grows from 0.5 to 5\%) and, also, to a relatively even distribution of images from iNaturalist, Observation.org, and Pl@ntNet (the corresponding fractions are 31, 29, and 35\%).

\subsection{Model}
\label{sec:models}

The process of deriving the Floralens model using GAMLV is illustrated in Figure~\ref{fig:AMLV-pipeline}. Overall, it comprises three stages: (1) preparing the dataset for training; (2) training the model, and (3) deploying the model onto a cloud server or (using a suitable format) onto edge devices.
GAMLV essentially requires the user to focus on the dataset preparation (1), given that training (2) and deployment (3) merely require simple high-level options by the user and are otherwise automated~\cite{Bisong2019,automl-vision-docs}.
The interaction with GAMLV can be conducted via a browser with a simple user interface, as we illustrate partially in this section (cf. Figure~\ref{fig:AMLV-train}), or programmatically using Google Cloud APIs (e.g., in Python). 

\begin{figure*}[b!]
\centering
\includegraphics[width=\linewidth]{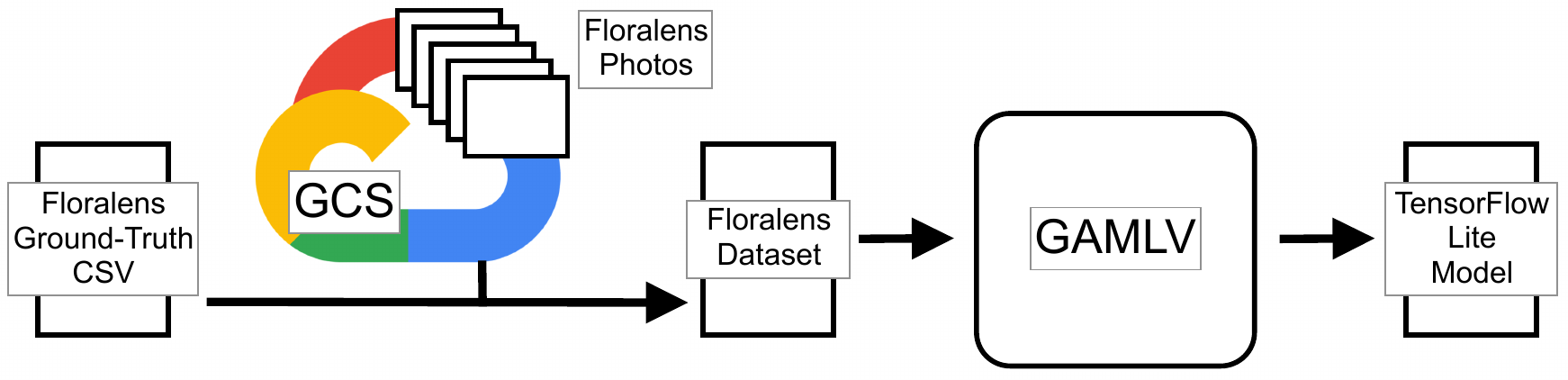}
\caption{Model derivation using GAMLV.}
\label{fig:AMLV-pipeline}
\end{figure*}

The first step requires the user to load the dataset images onto a storage bucket, in this case, provided by the Google Cloud Storage service (GCS), along with a simple CSV file. The latter lists the GCS image URIs, their respective ground truth labels (the name of the species in the image), and whether they belong to the training, validation, or test subset.

We provided GAMLV with train, validation, and test splits over the Floralens dataset, corresponding to fractions of 80\%, 10\%, and 10\%. As usual, the train split is used to adjust CNN parameters during the training process in an iterative feedback loop. The validation split is used to measure the progress and convergence of that training process. Finally, the test split is used for evaluating the model after training. The splits, with the image counts and data source provenance detailed in Table~\ref{tab:ds-split}, were obtained from a random selection of images for each species. Since the selection process is random and given the volume of images involved, the overall fraction of images of each data source in each split closely matches that of the overall dataset.

\begin{table}[t!]
    \caption{Train, validation, and test splits over the Floralens dataset.}
    \label{tab:ds-split}
    \centering
    {\small
    \begin{tabular}{@{}lr@{ }r@{ }r@{ }r@{ }r@{ }r@{ }r@{ }r@{ }r@{ }r@{ }r@{}}
\toprule
Data source & Train & & ~~Valid. & & ~~Test & & ~~Total & \\
\midrule
FloraOn         & 12,175 & & 1,466 & & 1,550 & & 15,191 & ( 5\%)\\ 
iNaturalist     & 72,174& & 8,924 & & 9,029 & & 90,127 & (31\%)  \\
Observation.org & 68,614 & & 8,603 & & 8,529 & &  85,746 & (29\%) \\
Pl@ntNet        & 81,918 & & 10,367 & & 10,252 & & 102,537 & (35\%)\\
\midrule
All             & 234,881 & (80\%) & 29,360 & (10\%)  & 29,360 & (10\%) & 293,601 & (100\%)\\
\bottomrule
\end{tabular}
    }
\end{table}

In~\cite{filgueiras2022}, we considered other strategies for defining these splits. In particular, we explored approaches that prefer specific data sources for the validation/test splits. We found that a random split, besides preserving a roughly similar fraction of images per data source in each split, results in models with better performance 
(contrast the results in Section~\ref{sec:results} with those in~\cite{filgueiras2022}). In any case, prioritizing particular data sources (e.g., FloraOn or Pl@ntNet) for validation/test splits over others had little impact on model performance. 

\begin{figure*}[ht]
\centering
 \begin{subfigure}[t]{0.46\textwidth}
        \centering
        \includegraphics[width=\textwidth]{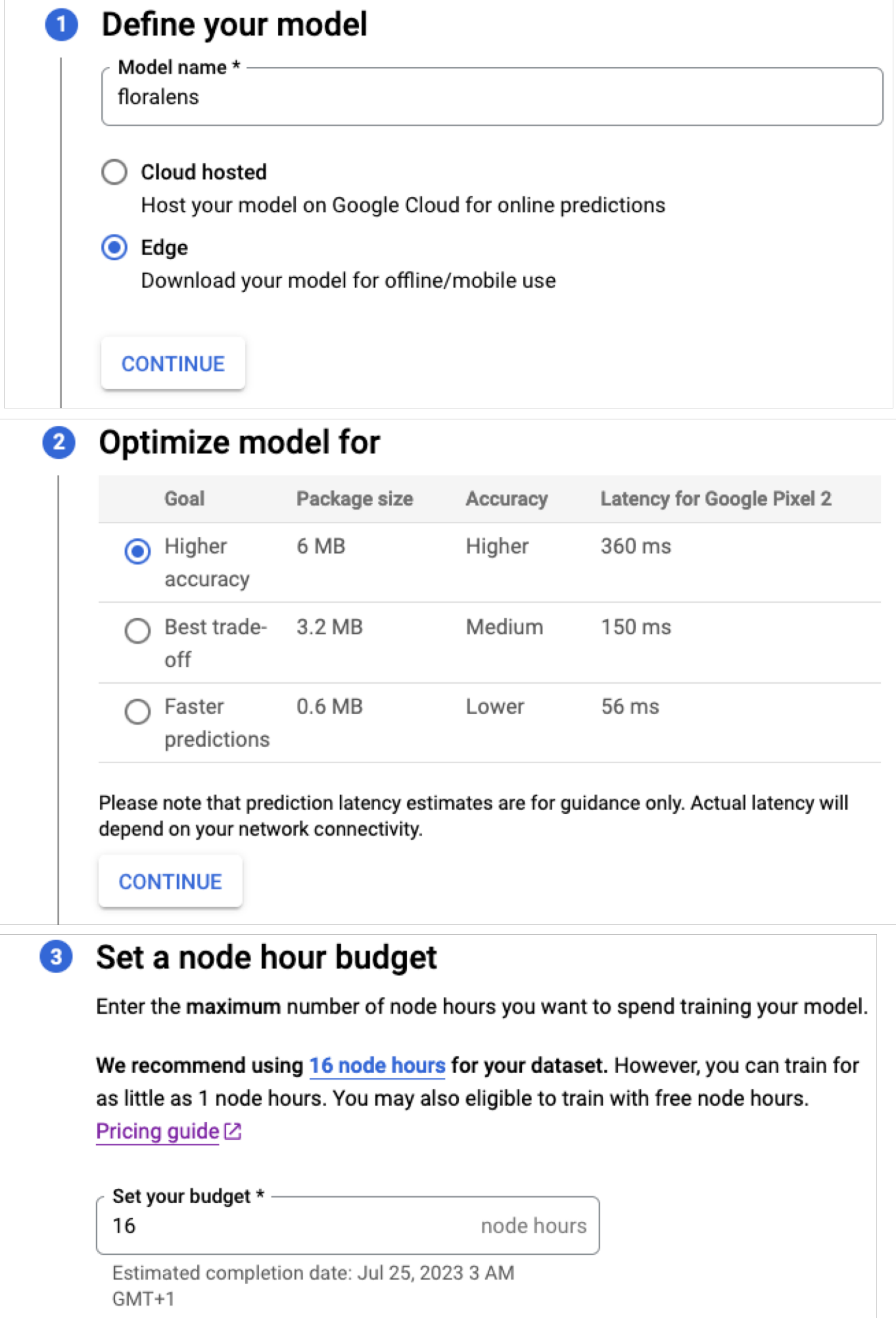}
        \caption{Training parameters.}
        \label{fig:amlv-training}
    \end{subfigure}
 \begin{subfigure}[t]{0.51\textwidth}
        \centering
        \includegraphics[width=\textwidth]{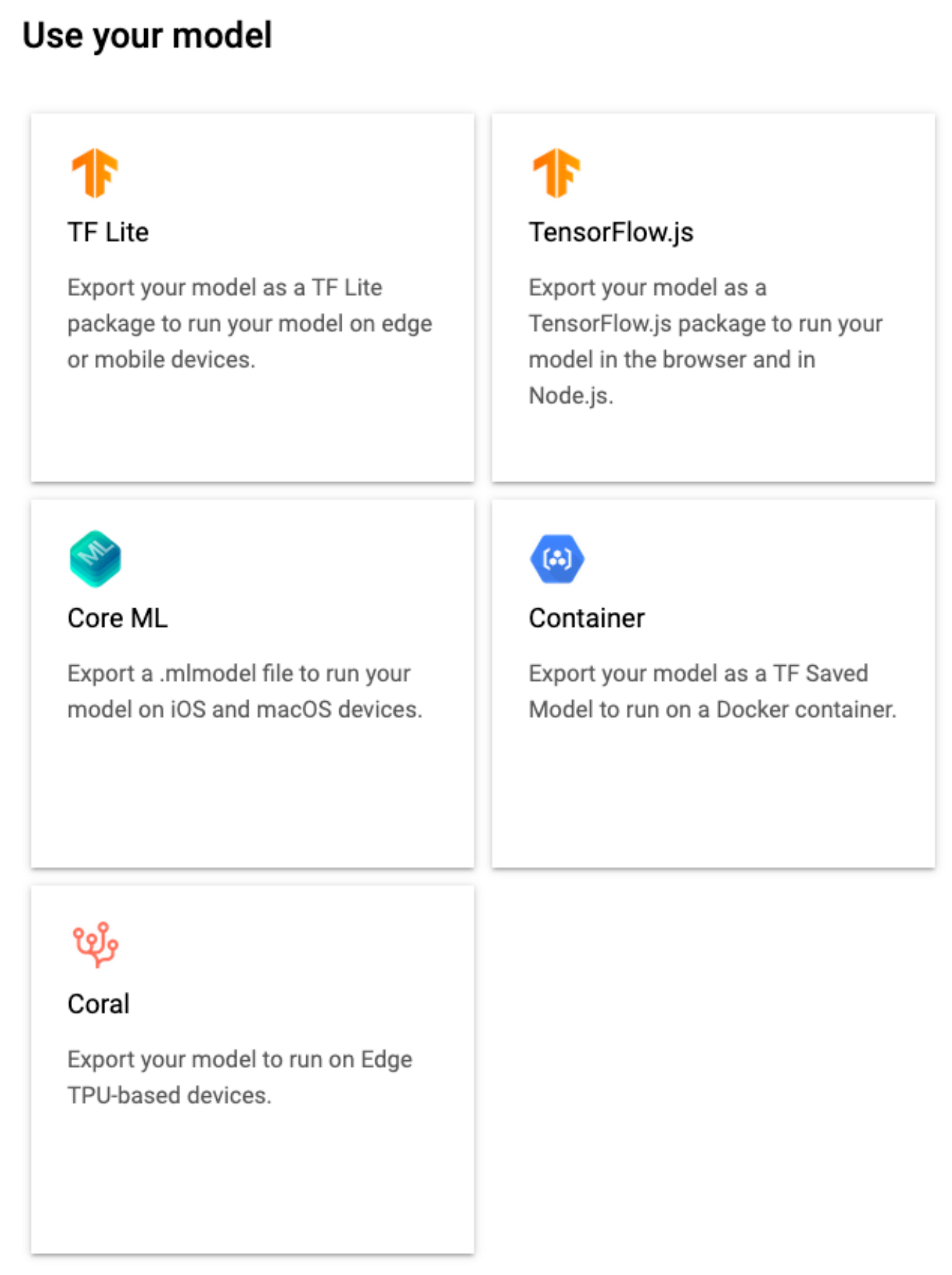}
        \caption{Deployment options.}
        \label{fig:amlv-deploy}
    \end{subfigure}
\caption{GAMLV interface for model training and deployment.}
\label{fig:AMLV-train}
\end{figure*}

Once the dataset is imported onto AutoML, the user selects high-level choices for the type of model to be generated and the maximum training time, as illustrated in Figure~\ref{fig:amlv-training}. Since we wish to use the model within web or mobile applications (cf. Section~\ref{sec:artifacts}), rather than deploying it on a Google Cloud server, we select the ``Edge'' model option. We also toggle the option for a model that favors accuracy over latency among the three available choices. The maximum training time is specified as a node-hours budget, where nodes are virtual machines used during training. Once this configuration is done, training may proceed.

GAMLV required 4 node hours to complete the training of the CNN with the Floralens dataset. The service operates as a black box, providing no insight into the internal training process or infrastructure configuration. For instance, no exact details or configuration options are provided for the training infrastructure (e.g., virtual machines, GPUs, or TPUs), and it is not possible to track details regarding the training process (e.g., how the model converges over time).

Once training is completed, a model can be deployed in one of several formats amenable for integration with a local application (Figure~\ref{fig:amlv-deploy}). The formats include the standard SavedModel format used by TensorFlow,  but also others like TF Lite~\cite{tf-lite}, a lightweight TensorFlow format for use in resource-constrained hosts (e.g., mobile and embedded devices), or TFJS~\cite{tfjs}, for use in web browsers or JavaScript programs. We used the TF Lite and TFJS variants in the software artifacts described in Section~\ref{sec:artifacts}. 

\begin{figure}[h!]
    \centering
    \begin{subfigure}[b]{0.3\textwidth}
        \centering
        \includegraphics[height=7cm]{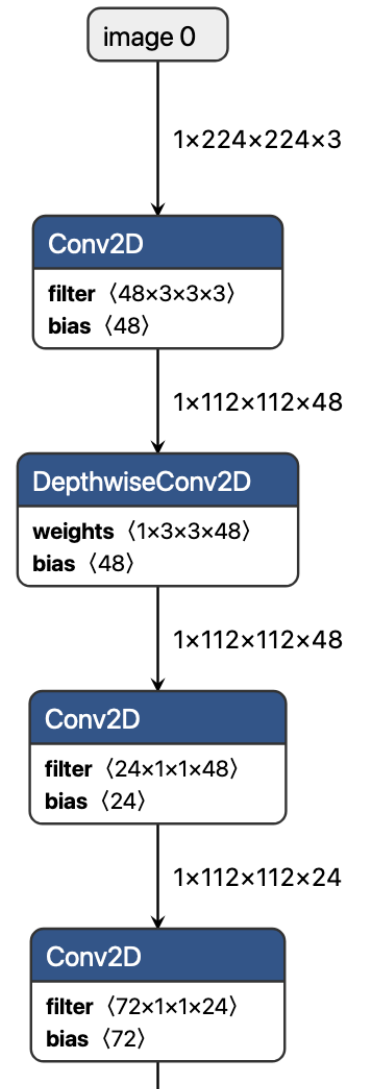}
        \caption{Initial (including input layer).}
        \label{fig:cnnp1}
    \end{subfigure}
    \begin{subfigure}[b]{0.3\textwidth}
        \centering
        \includegraphics[height=7cm]{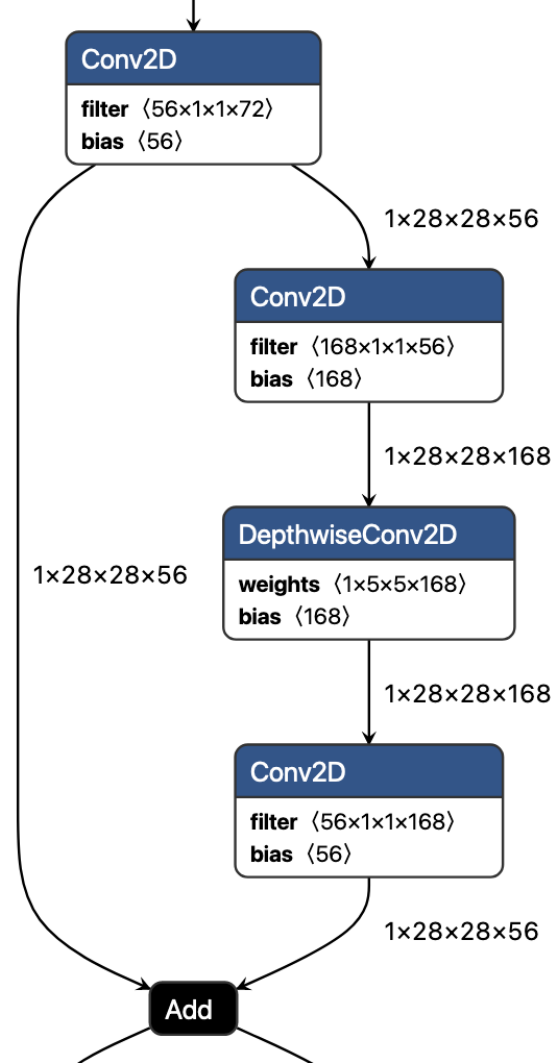}
        \caption{Intermediate.}
        \label{fig:cnnp2}
    \end{subfigure}
    \begin{subfigure}[b]{0.3\textwidth}
        \centering
        \includegraphics[height=7cm]{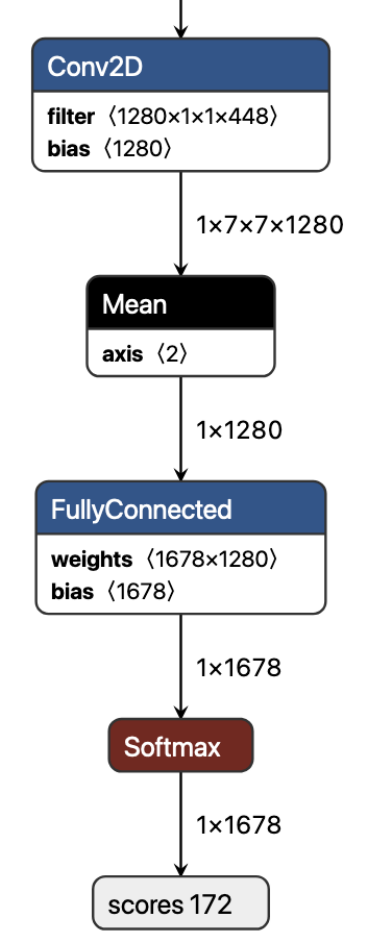}
        \caption{Final (including output layer).}
        \label{fig:cnnp3}
    \end{subfigure}
    \caption{Layers of the CNN model used for Floralens (fragment).} 
    \label{fig:automl_cnn}
\end{figure}

The model obtained by GAMLV is a 65-layer deep CNN, with the structure partially illustrated in Figure~\ref{fig:automl_cnn} for the TF Lite version, in terms
of the input/initial layers  (a), intermediate layers (b), and final/output layers (c). Each box in the diagram represents a neuron type in the CNN. A neuron processes a multi-dimensional vector (a tensor) and outputs another tensor. Its internal weights are adjusted iteratively using backpropagation during training. CNNs utilize a particular family of functions called convolutions, which are especially suited for detecting image features (e.g., edges). In the simplest type of convolution, operating over 2D matrices, each position in the output vector, designated a feature map, is the dot product of a sliding window over the input matrix with a filter defined by the internal weights of the neuron. For details, see for instance Chapter 9 of~\cite{Goodfellow-et-al-2016}.

The TF Lite version differs from the standard TensorFlow model only in terms of post-training optimizations such as quantization (the conversion of floating-point weights to an integer scale). This allows a faster interpretation of the model with negligible degradation in accuracy~\cite{tf-lite}. As shown in Figure~\ref{fig:automl_cnn}, the input layer takes a $224\times224\times3$ tensor, corresponding to a $224\times224$  (typically resized to conform with the dimensions of the CNN input) image with 3 RGB channels, with 8-bit values per color channel. The intermediate layers use several types of convolutions in a repeating pattern. The final layers include the derivation of a 1,280-feature map (a vector summarizing all the captured image features~\cite{Goodfellow-et-al-2016}). This feature map is then passed to a fully connected softmax activation function, which produces the final classification vector with probabilities for the labels, 1,678 of them in line with the number of species covered.
In this context, fully connected means that every value of the feature map, combined with the internal weights, is taken into account to calculate the value of each output position. The softmax activation function is used to produce the neural network's final output, which, in this case, is a probability distribution~\cite{Goodfellow-et-al-2016}.
As expected for a probability distribution, the sum of all the probabilities in the output vector equals 1. Note that we always get such a vector, even for images outside the domain in question (in our case, the native Portuguese flora).  If the model is certain
of a particular species, we get a high probability value for that species and low values for all the others. The model will be less certain, for instance, when the probabilities assigned to the most highly ranked species are close (e.g., two or more similar species). Finally, when the model fails to make any meaningful identification, we get low probability values for all species.

The CNN architectures used are from the MnasNet family~\cite{mnas-net}, developed with mobile and embedded devices in mind. The high-level choice between models offered by GAMLV, depicted in Figure~\ref{fig:amlv-training}, corresponds to three different MnasNet instantiations that do not differ in structure, just in the density of connections between layers and the number of internal weights.

\subsection{Evaluation Metrics}

The performance of the Floralens model was evaluated across several test datasets of observations using the following metrics. Precision is the ratio of true positives (TP) relative to the total number of positives (TP $+$ FP). A positive (identification) occurs when the classification score (a probability) returned by the model equals or exceeds a confidence level set as the threshold for analysis. Recall is the ratio of true positives relative to the total number of true examples (TP $+$ FN). The confidence level is relevant for practical uses of a model. A low confidence level implies tolerance for false positives in the intended application and results in lower precision. On the other hand, a high confidence level will result in more false negatives, hence lower recall.
\begin{gather*}
\text{Precision} = \frac{\text{TP}}{\text{TP}+\text{FP}} \qquad 
\text{Recall} = \frac{\text{TP}}{\text{TP}+\text{FN}}
\end{gather*}
Top-1 is the fraction of test images that the model correctly classified by the label with rank 1 (the highest-scoring label). Top-5 is similar to Top-1 but also includes test images with a rank lower than or equal to 5 (the 5 highest-scoring labels). We also computed a variant of the Mean Reciprocal Rank (\mrrs{}) restricted to test images with ranks less than or equal to 5. These are defined as follows:
\begin{gather*}
Q(r_l) = \{ t \in  T  \mid  \mathit{rank}(t) \leq r_l \}   \\ \\
\text{Top-1}  = \frac{\lvert Q(1) \rvert}{\lvert T \rvert} \qquad \text{Top-5}  = \frac{\lvert Q(5) \rvert}{\lvert T \rvert} \qquad 
\text{MRR} = \frac{1}{\lvert T \rvert} \sum_{t \in Q(5)}  \frac{1}{\mathit{rank}(t)}   
\end{gather*}
where $T$ is, as above, the set of all test observations ($\lvert T \rvert=$ 29,360 as given in Table~\ref{tab:ds-split}), $\mathit{rank}(t)$ is the rank of the ground truth label returned by the model for the test observation $t$, and $Q(r_l)$ is the subset of $T$ that contains test observations with rank less or equal to a limit $r_l$.

\section{Results}
\label{sec:results}

In this section, we present the results of multiple experiments conducted to characterize the performance of the Floralens model and compare it to state-of-the-art models. In the following subsections, higher values indicate better model performance.

\subsection{Baseline Results}
\label{results:baseline}

Figure~\ref{fig:precision-recall} shows the macro-averaged precision and recall of the Floralens model on the base test dataset (henceforth FLTS, cf.\ Table~\ref{tab:ds-split}). Figure~\ref{fig:precision-recalla} reports an area under the curve (AUC) of $0.72$ (maximum = $1.0$), which corresponds to the average precision metric.  Figure~\ref{fig:precision-recallb} shows that precision and recall are both near  $0.7$ for a confidence threshold of $0.2$. When the confidence threshold is set to $0.5$, precision increases to $0.85$, while recall decreases to $0.53$. Overall, the results indicate that the Floralens model has good predictive power.

\begin{figure}[h!]
\centering
\begin{subfigure}[b]{0.49\textwidth}
  \includegraphics[width=\textwidth]{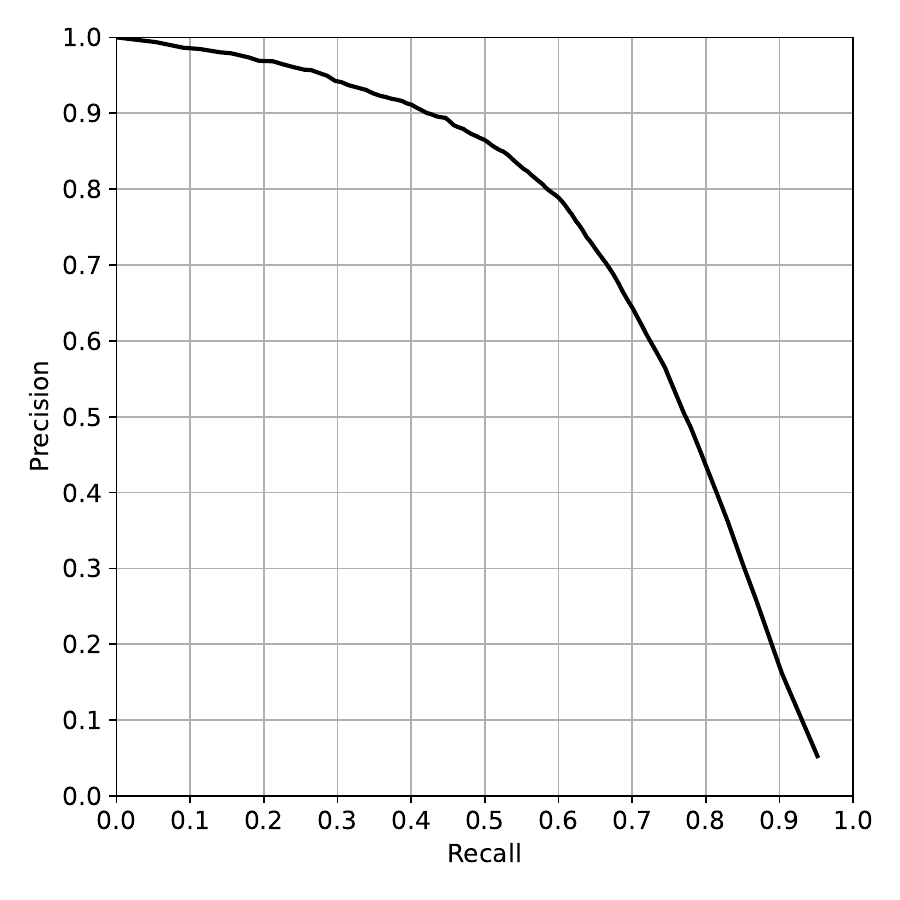}
  \caption{Precision-Recall curve.\label{fig:precision-recalla}}
\end{subfigure}
\begin{subfigure}[b]{0.49\textwidth}
 \includegraphics[width=\textwidth]{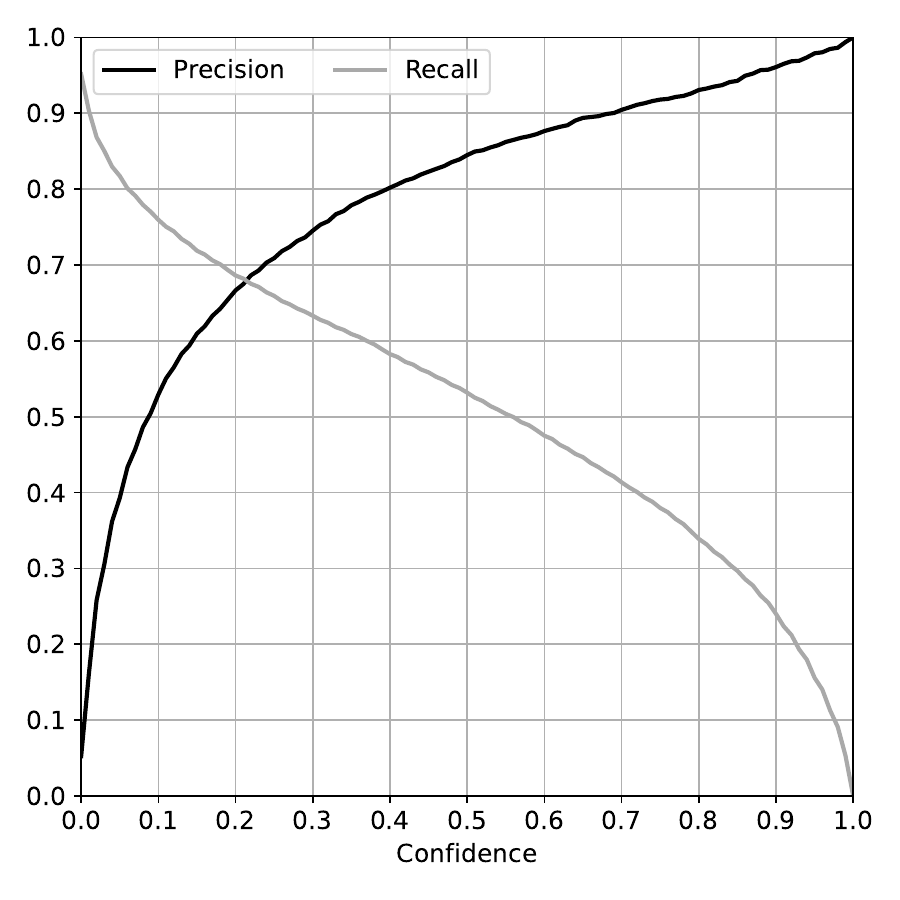}
  \caption{Precision and recall vs.\ confidence.\label{fig:precision-recallb}}
\end{subfigure}
\caption{FLTS results: precision and recall.\label{fig:precision-recall}}
\end{figure}

\begin{figure}[h!]
\centering
\begin{subfigure}{0.8\linewidth}
  \includegraphics[width=\linewidth]{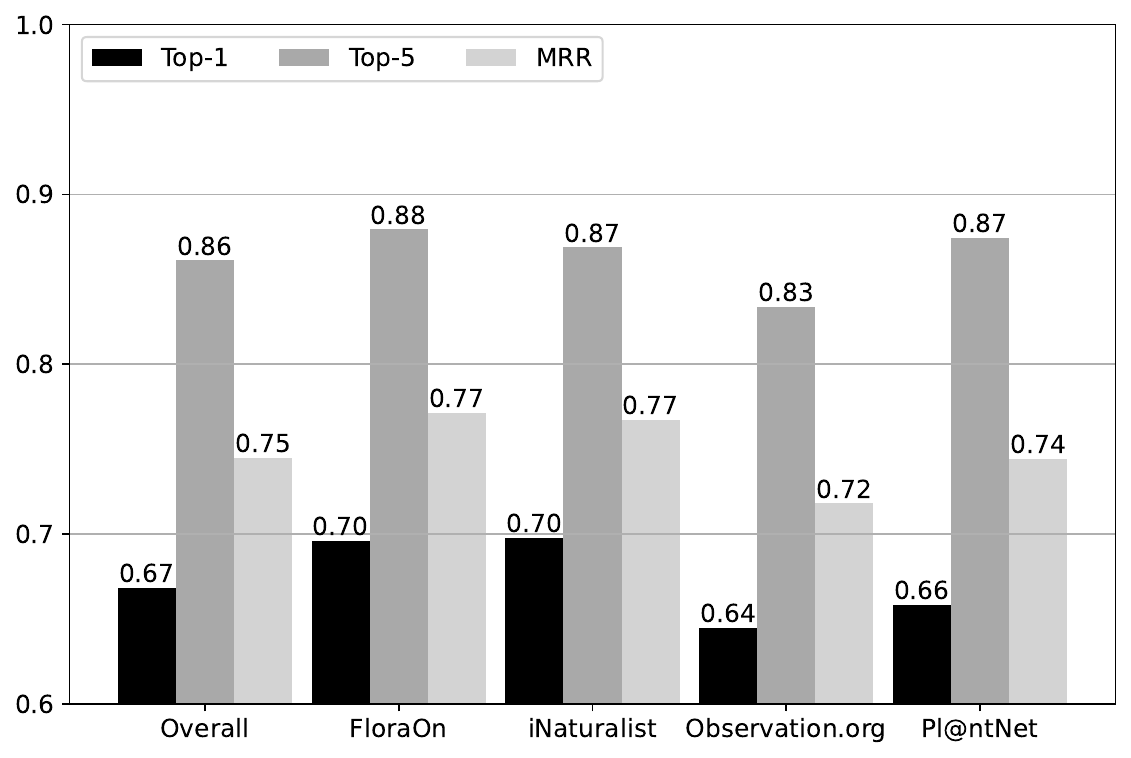}
  \subcaption{Top-1, Top-5, and MRR results: overall and per data source.\label{fig:floralens_baseline_results}}
\end{subfigure}
\begin{subfigure}{0.6\linewidth}
  \includegraphics[width=\linewidth]{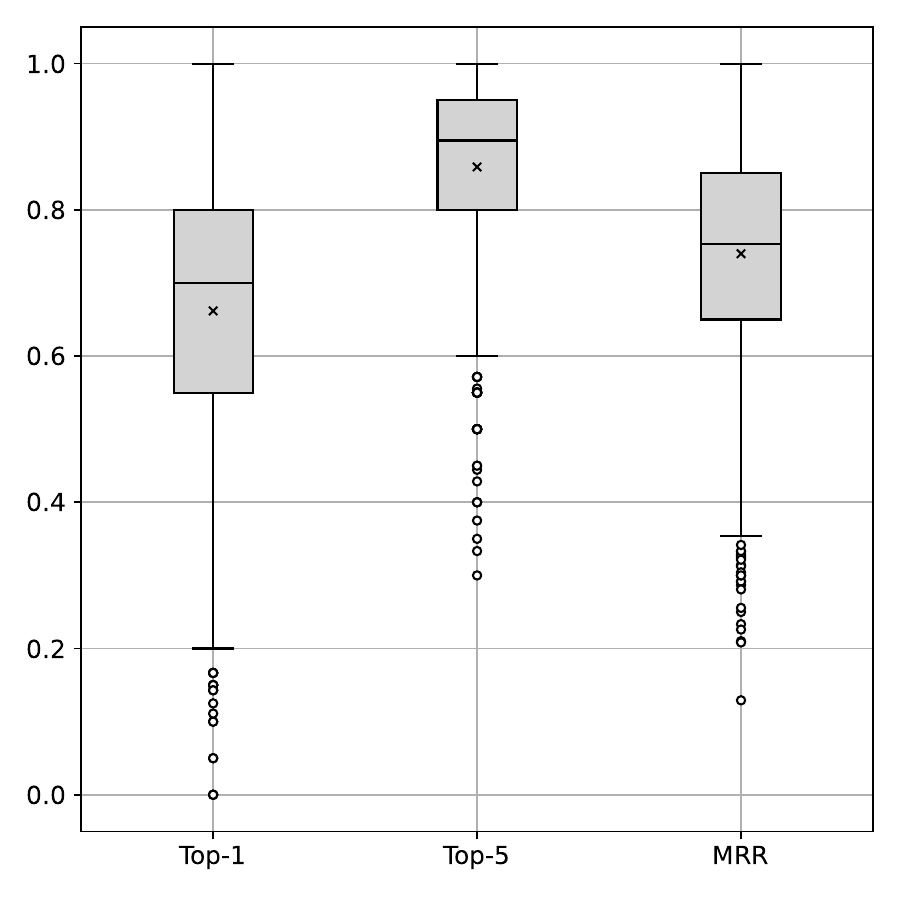}
  \subcaption{Distribution of results per species.\label{fig:floralens_baseline_results-boxplots}}
\end{subfigure}
\caption{Floralens baseline results for the FLTS dataset. \label{fig:floralens_baseline_results-top1_top5_mrr}}
\end{figure}

\begin{figure}[h!]
\centering
    \begin{subfigure}{0.8\linewidth}
        \centering
        \includegraphics[width=\textwidth]{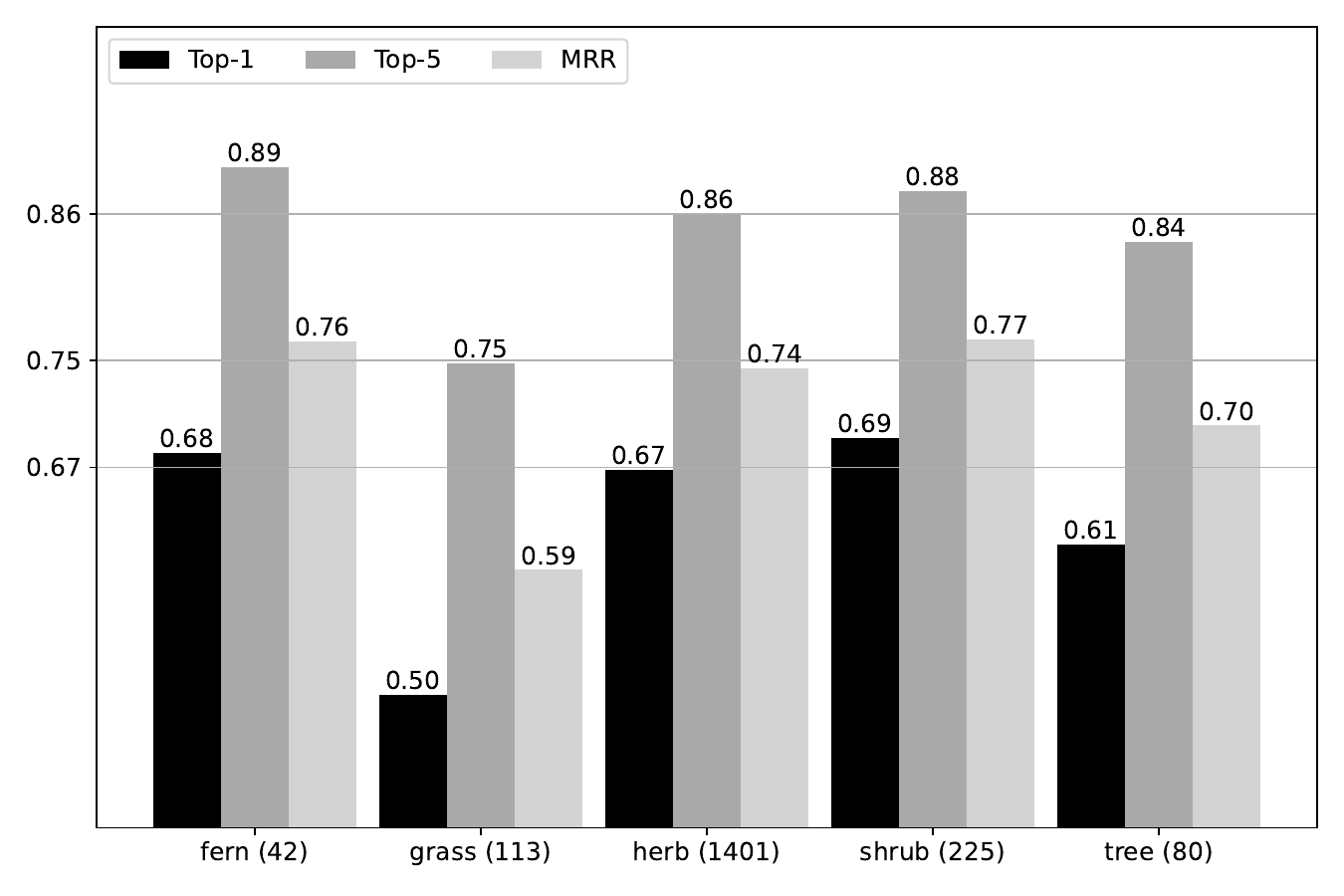}
        \subcaption{Top-1, Top-5, and MRR results as a function of growth form.}
        \label{fig:gf_results}
    \end{subfigure} \\
    \begin{subfigure}{0.8\linewidth}
        \centering
        \includegraphics[width=\textwidth]{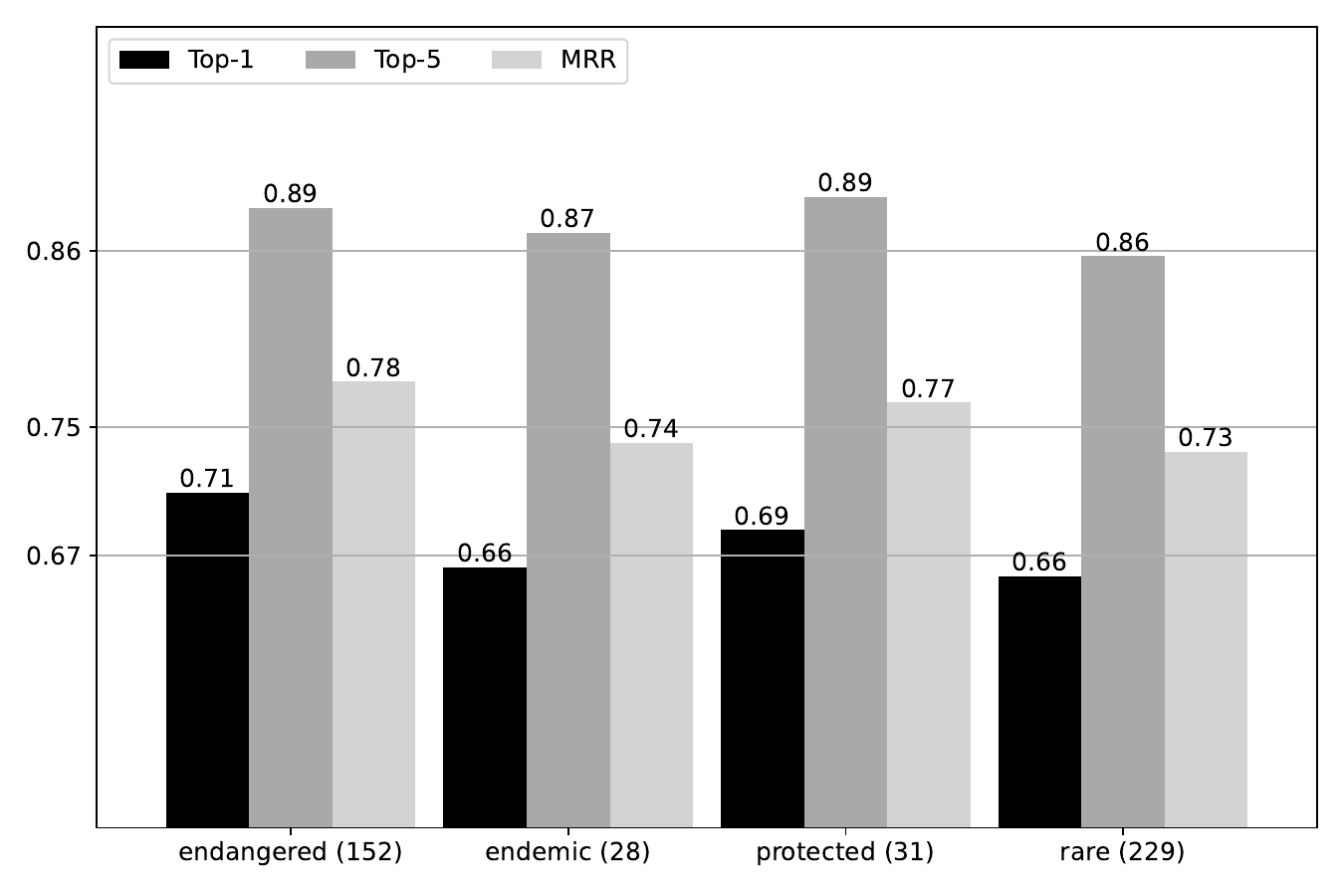}
         \subcaption{Top-1, Top-5, and MRR results for special categories.}
        \label{fig:cs_results}
    \end{subfigure} 
    \caption{Results according to species growth form and special categories.\label{fig:gf_and_cs_results}}
\end{figure}

Figure~\ref{fig:floralens_baseline_results-top1_top5_mrr} shows the Top-1, Top-5, and \mrrs{} results for FLTS in two plots. Figure~\ref{fig:floralens_baseline_results} shows the overall results together with the results per data source used in the construction of the dataset (FloraOn, iNaturalist, Observation.org, and Pl@ntNet). The boxplots in Figure~\ref{fig:floralens_baseline_results-boxplots} display the distribution of the metric results across species. 
Again, the overall results exhibit good predictive performance: $0.67$ for Top-1 (two-thirds of the FLTS images correctly classified), $0.86$ for Top-5, and $0.75$ for MRR. Performance is consistent across data sources, with differences not exceeding $0.06$ (Top-1 is $0.64$ for Observation.org and $0.70$ for both FloraOn and iNaturalist).
Regarding the distribution of results per species, the median values for Top-1, Top-5, and MRR are respectively $0.70$, $0.89$, and $0.75$. The first quartile values also support this performance: $75$\% of species achieve a Top-1 accuracy above $0.55$, while the first quartile for Top-5 and MRR is $0.80$ and $0.65$, respectively.

Figure~\ref{fig:gf_and_cs_results} presents two types of complementary results. The plots detail the Top-1, Top-5, and MRR grouped by (\ref{fig:gf_results}) species growth form and (\ref{fig:cs_results}) the special categories of species that are endangered, endemic, protected, or rare. In both cases, the FloraOn site~\cite{flora-on} is the reference for the presented subsets of species. The site uses the IUCN Red List taxonomy~\cite{iucn}
for endangered species and Clive Stace's classification~\cite{stace2010new} for rare species.
 The number of species in each class is indicated in parentheses in the figures (e.g., $42$ ferns). The y-axis reference ticks correspond to the overall FLTS values for the Top-1 ($0.67$), MRR ($0.75$), and Top-5 ($0.86$) metrics. Grasses show the lowest accuracy, likely reflecting the difficulty of visual distinction. Tree species also perform slightly below the overall FLTS average. For the special categories, the results remain close to the overall FLTS values. The results are encouraging, particularly for endemic species, which are often challenging for models to identify~\cite{mindyourapp}.

\subsection{PlantCLEF and Wikipedia Datasets}

We analyze two additional datasets: a random sample of 10,000 labeled images from the PlantCLEF'22-23~\cite{plantclef2022} competition,
and a sample of close to 1,500 images from Wikipedia. 
The PlantCLEF data used here represent a small subset of the full ``trusted" training set~\cite{planclef_trusted_training_set}, which contains about $2.9$ million images across 80,000 plant species. The repository is trusted as the image labels were obtained from academic sources or collaborative platforms like Pl@ntNet or iNaturalist. Our subset was constructed by first filtering out species not covered by the Floralens model, providing us with data for 1,593 (out of 1,678) species. From these data, we then randomly sampled 10,000 images.

As for the Wikipedia dataset, the images were identified through the Wikimedia REST API~\cite{wikimedia_rest_api}. For each species in the Floralens domain, we used the species name as the keyword for a REST API search. The search result typically yields a reference to an image stored at Wikipedia, which we then considered for addition to the dataset. 
After collecting the images, we filtered out illustrations, herbarium specimens, and duplicates linked to multiple species. Duplicates often occur when the search returns an image of a different species from the same genus—for instance, when the target species lacks its own Wikipedia page. Wikipedia identifications tend to be less reliable, as they result from a crowd-sourced effort with no specific directives for species validation.
Through this process, we obtained a dataset of 1,351 images for an equal number of species (one per species).  

Figure~\ref{fig:sgf_species} shows the results of the Floralens model for these datasets evaluated in terms of the Top-1, Top-5, and \mrrs{} metrics for species prediction. We also recall the overall FLTS results (from Figure~\ref{fig:floralens_baseline_results}) for easy comparison. The results for the PlantCLEF and Wikipedia datasets are marginally lower than those for FLTS ($0.02$ to $0.03$ in all metrics), suggesting that the Floralens model performs well with datasets other than our base test suite.

\subsection{Genus and Family Identification Results}

\begin{figure}[t!]
\centering
    \begin{subfigure}{0.49\linewidth}
        \centering
        \includegraphics[width=\textwidth]{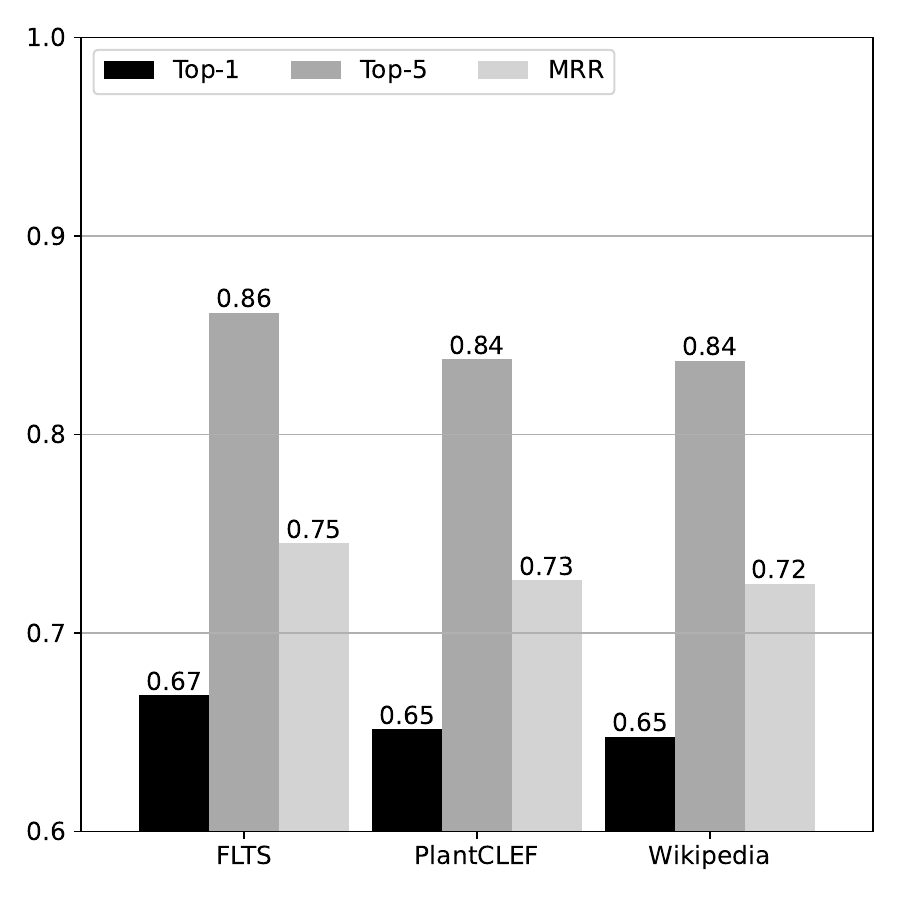}
        \subcaption{Top-1, Top-5, and MRR for species and per test dataset.}
        \label{fig:sgf_species}
    \end{subfigure} \\
    \begin{subfigure}{0.49\linewidth}
        \centering
        \includegraphics[width=\textwidth]{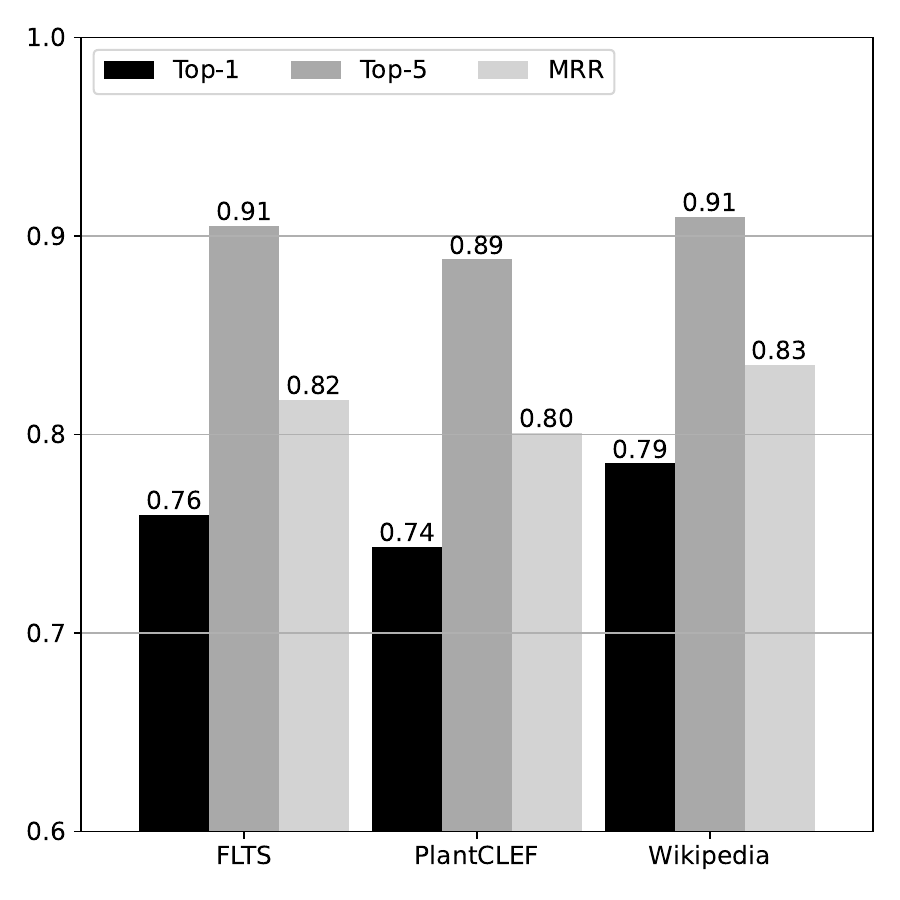}
        \subcaption{Top-1, Top-5, and MRR for genus and per test dataset.}
    \label{fig:sgf_genus}
    \end{subfigure} 
    \begin{subfigure}{0.49\linewidth}
        \centering
       \includegraphics[width=\textwidth]{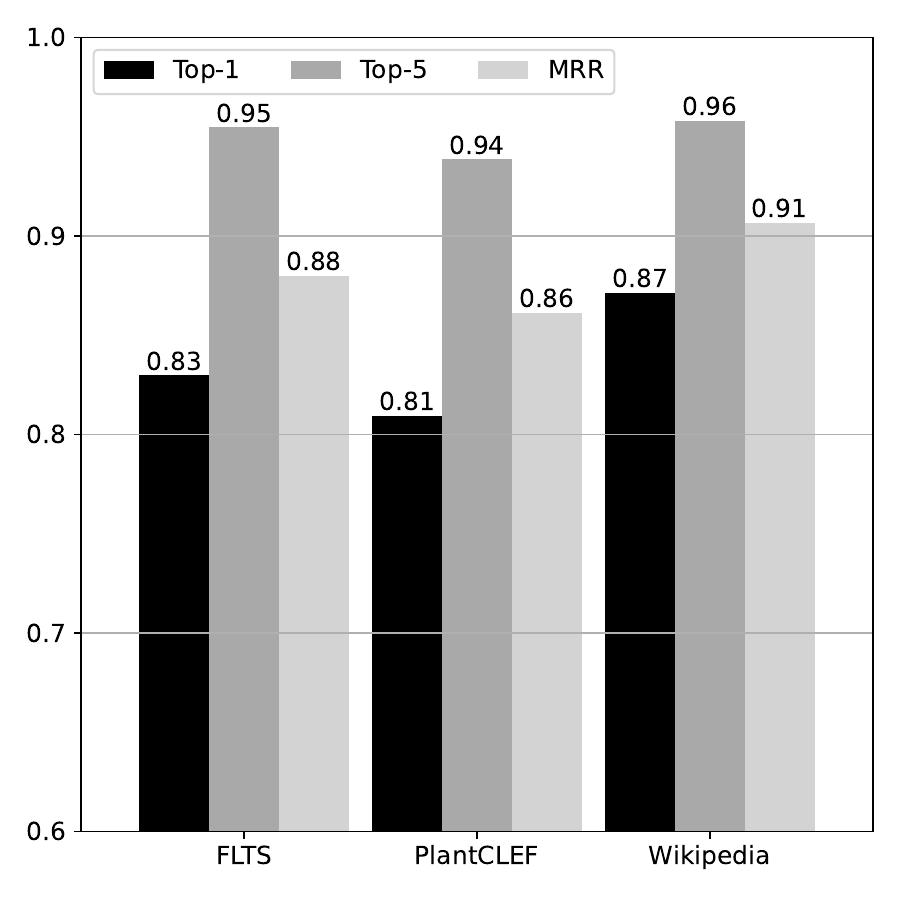}
        \subcaption{Top-1, Top-5, and MRR for family and per test dataset.}
    \label{fig:sgf_family}
    \end{subfigure}
\caption{Species, genus, and family identification results for all test datasets.\label{fig:genus_family}}
\end{figure}

We now report results also at the genus and family levels to assess the robustness and generalizability of the Floralens model.

Figures~\ref{fig:sgf_genus} and~\ref{fig:sgf_family} show the results of genus and family identification for all datasets. The classification score for a genus or family is obtained by aggregating the scores of all species belonging to that taxon. In previous work~\cite{filgueiras2022}, we trained a separate model for genus identification but found its performance to be nearly identical to that achieved using this aggregation approach. Figure~\ref{fig:sgf_species} shows that genus‑level Top‑1 accuracy exceeds species‑level by $0.09$–$0.14$, Top‑5 by $0.05$–$0.07$, and MRR by $0.07$–$0.11$. In turn, the family results outperform genus results for Top-1 by $0.07$–$0.08$, Top-5 by $0.04$–$0.05$, and MRR by $0.06$–$0.08$. As with species, the results for genus and family are homogeneous across all datasets. The most significant variation is observed in the genus results of the Wikipedia dataset, with the Top-1 result being $0.14$ higher than the species‑level Top‑1 result. This might be explained by the fact that even when an image on a species page is incorrectly labeled, Wikipedia usually manages to provide one for a plant of the same genus and, naturally, family.

\subsection{Classification using Multiple Images}
\label{results:multiple}

It is often the case that an observation of a plant in the wild has multiple images associated with it to enable better identification. In this section, we take advantage of this fact and compute an aggregate score for each such observation based on the scores returned by Floralens for each image of a given specimen. We apply this operation to all 
observations with two or more associated images in the FLTS.

\begin{figure}[h!]
\centering
\includegraphics[width=0.75\linewidth]{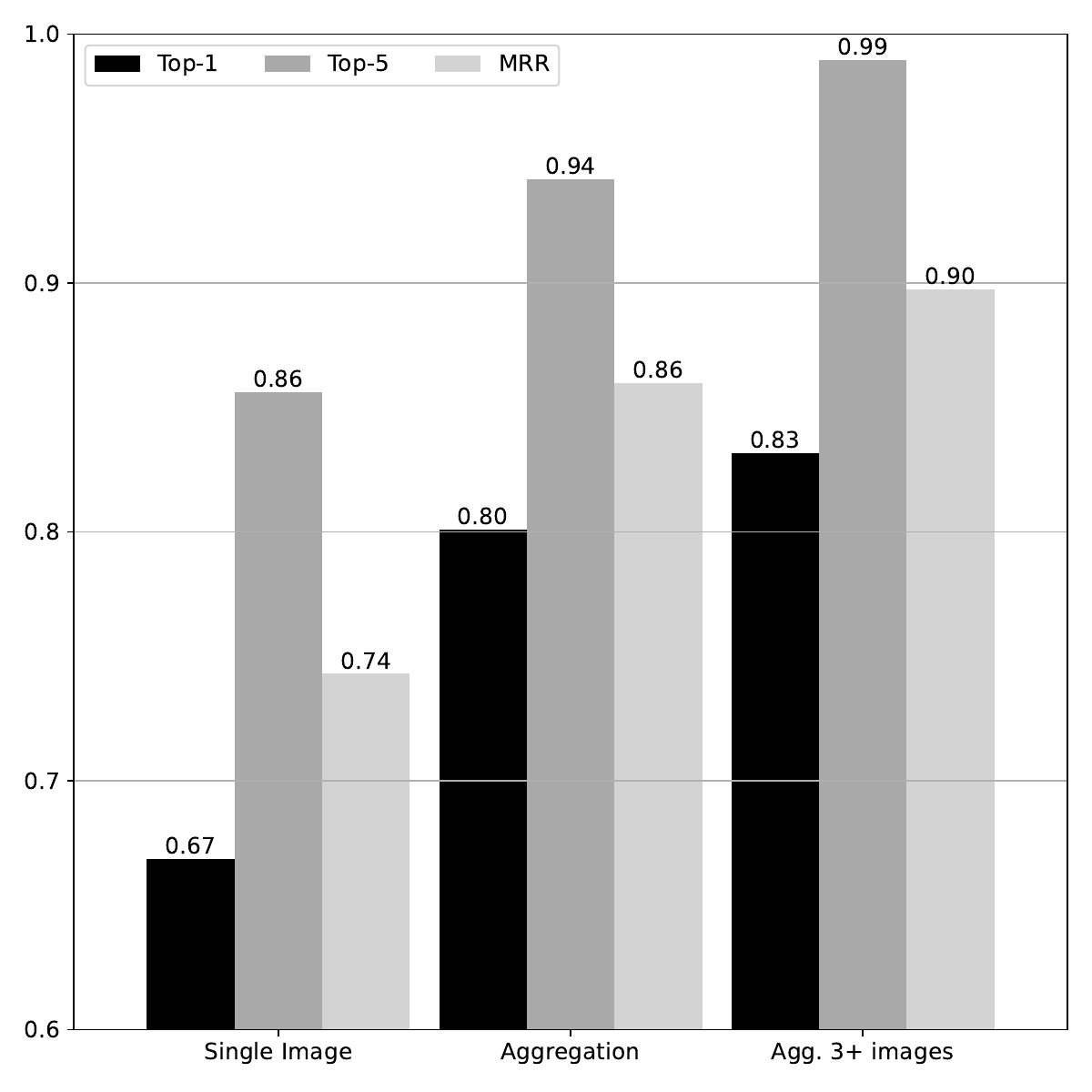}
\caption{Top-1, Top-5, and MRR results for classification using multiple images.\label{fig:multiple_image_results}}
\end{figure}

Among the 1,185 multi‑image observations (totaling 2,480 images), $91$\% contain exactly two images. The remaining 95 observations include three ($82$), four ($12$), or six ($1$) images. These observations are grouped by the GBIF identifier of images in the iNaturalist, Observation.org, and Pl@ntNet datasets. For FloraOn, we lacked metadata linking images to the same observation, so we were not able to group them. Our analysis was not extended to the PlantCLEF dataset for the same reason. In the Wikipedia dataset, there is only one image per species.

For the 1,185 observations of single specimens, we calculated an aggregate classification score as the average of the scores of the individual images. Figure~\ref{fig:multiple_image_results} compares the standalone image classification (the baseline FLTS results) vs.\ the aggregate classification of observations in terms of Top-1, Top-5, and \mrrs{}. We observe clear score improvements: $+0.13$ for Top-1 ($0.67$ to $0.80$), $+0.08$ for Top-5 ($0.86$ to $0.94$), and $+0.12$ for \mrrs{} ($0.74$ to $0.86$). The figure also depicts (rightmost) the results for the (relatively few) $95$ observations ($9$\% of the cases) with $3$ or more images exhibiting even more pronounced gains.

\subsection{Classification using Geographical Location}
\label{results:location}

\begin{figure}[h!]
\centering
\begin{subfigure}{0.49\linewidth}
        \centering
        \includegraphics[width=\textwidth]{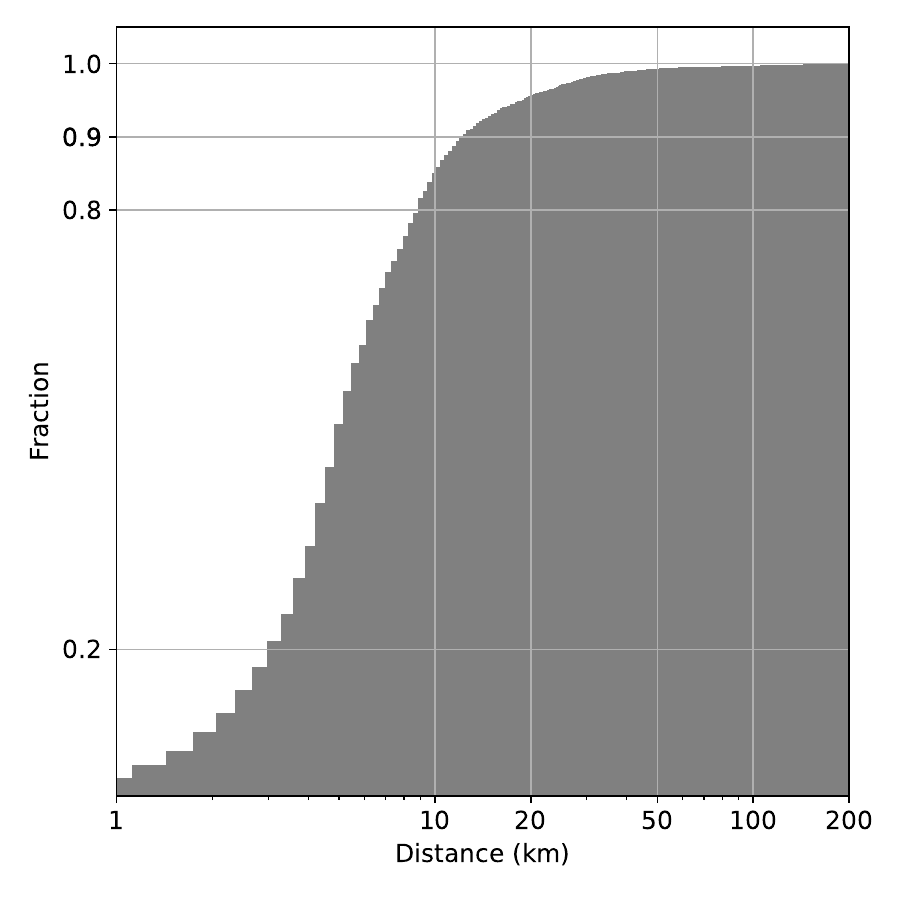}
        \subcaption{CDF plot for the shortest distance between GBIF coordinates and FloraOn grid records.}
        \label{fig:floraon_gbif_distance_cdf}
    \end{subfigure}
    \\
    \begin{subfigure}{0.49\linewidth}
        \centering
       \includegraphics[width=\textwidth]{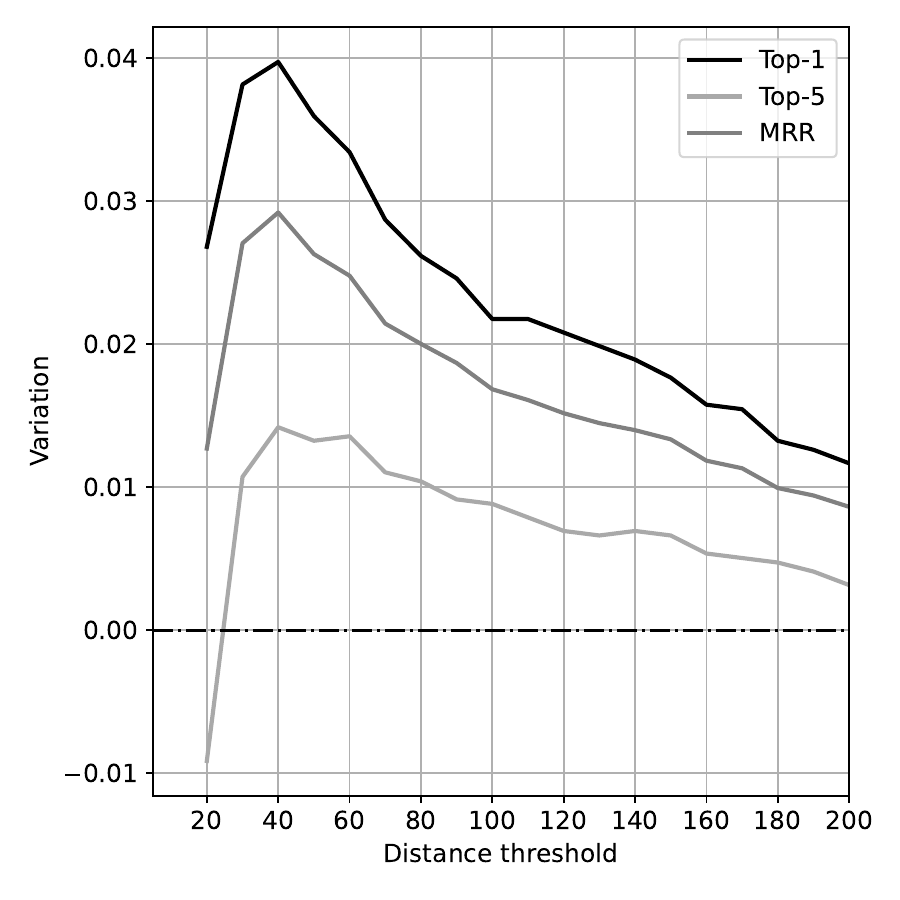}
        \subcaption{Top-1, Top-5, and MRR variations for different values of $D$ parameter of the geographical location filter.}
        \label{fig:geo_filter_plot}
    \end{subfigure}  
     \begin{subfigure}{0.49\linewidth}
        \centering
       \includegraphics[width=\textwidth]{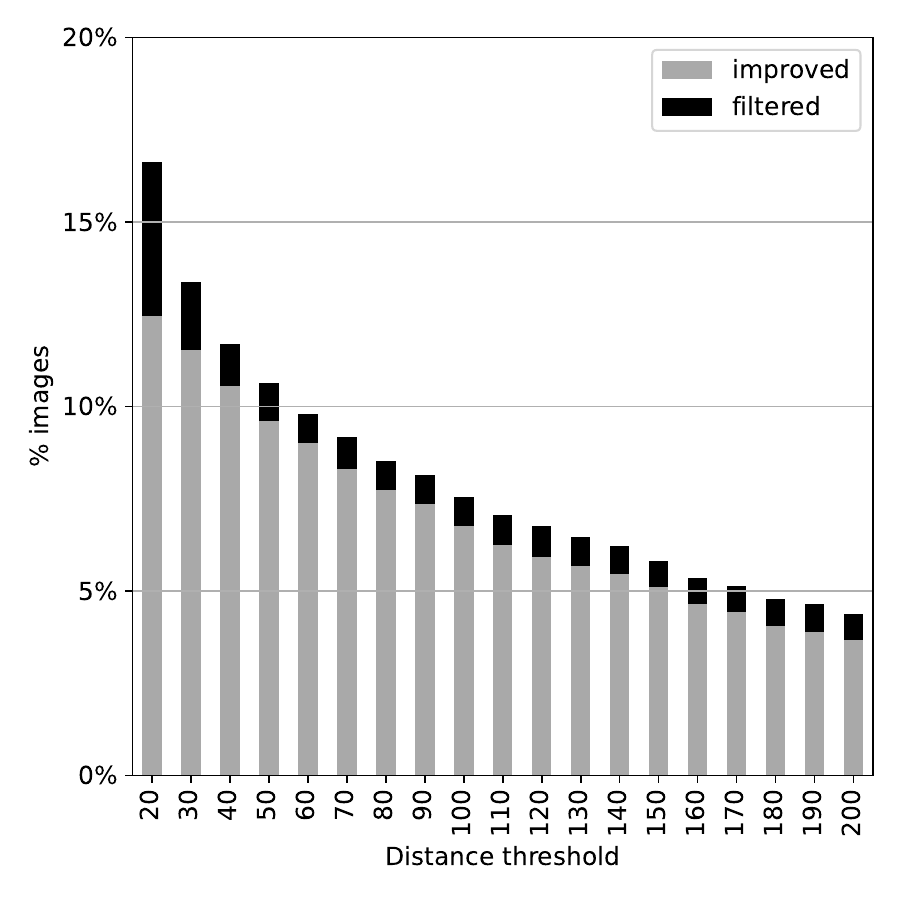}
        \subcaption{Percentage of images with altered ground truth ranks for different values of the $D$ parameter of the geographical location filter.}
        \label{fig:geo_filter_plot_changes}
    \end{subfigure} 
\caption{Results for the parametric geographic data filter.}\label{fig:geographic}
\end{figure}

Some automated classification platforms utilize the geographical location associated with observations to improve the accuracy of the results. For instance, Pl@ntNet provides a parameter for the geographical zone (e.g., Southwestern Europe) and filters out species that do not occur in that region from the classification results, regardless of the output of the image classification model. We analyzed the impact of this strategy by making use of two items of information: (1) detailed FloraOn records of the geographical location of observations in Continental Portugal (excluding the Azores and Madeira archipelagos), and (2) the geographical location associated with FLTS images obtained from GBIF.

FloraOn geographic distribution data are based on the Military Grid Reference System (MGRS) grid with a  $10$~km$\times$$10$~km resolution. For any given species, it keeps a set of MGRS grid elements for which observations have been reported. We collected these data 
for almost all species covered by the Floralens dataset -- 1,654 out of 1,678 (98\%). Next, we identified all images related to observations in Continental Portugal in the FLTS through GBIF. These amount to 3,172 images covering 770 species. We assessed spatial concordance by computing, for each image, the shortest distance between its GBIF coordinates and the corresponding FloraOn grid cells for the ground-truth species. The cumulative distribution function (CDF) of the shortest distances is shown in Figure~\ref{fig:floraon_gbif_distance_cdf}. Note the logarithmic x-axis scale. Distances between GBIF coordinates and FloraOn grids show strong agreement: 95\% $\leq$ 18.9 km, 97\% $\leq$ 24.7 km, and 99\% $\leq$ 42.4 km.

For each test image, the model produced a ranked list of candidate species. We then measured the distance $d$ between the observation’s location and the nearest known occurrence of each candidate species, discarding those with $d > D$, where $D$ is the chosen threshold. We applied this geographical filter for varying values of the distance threshold~$D$ from 20 to 200~km with increments of 10~km. Note that the granularity of the MGRS grid used in FloraOn is 10~km, thus justifying the choice of 20~km as the lowest value for~$D$. To gauge the impact of the filter, we calculated the variations in the Top-1,  Top-5, and MRR metrics when compared to the respective values without applying the filter. With no filter, we obtain 0.74 for Top-1, 0.89 for Top-5, and 0.80 for MRR. The variations are shown in Figure~\ref{fig:geo_filter_plot}. As illustrated, we observe small positive variations suggesting that a geographical filter is beneficial, though its effect is modest. The greatest improvement occurs at $D=40$ km: Top‑1 +0.04, Top‑5 +0.02, and MRR +0.03. 

Figure~\ref{fig:geo_filter_plot_changes} shows the fraction of images whose classification results were affected by the geographical filter. The plot distinguishes the following two cases: (1)  the ground truth (species) rank is improved, and (2) the ground truth is filtered out from the results.
At $D=20$ km, 12\% of images show an improved ground‑truth rank, while 4\% have the ground truth filtered out. As $D$ increases, the fraction of affected images decreases to less than 5\%. Note that the filter for $D=20$~km is too aggressive, incorrectly removing species from the rankings and hindering any gains from improving the ground truth's ranking position. This effect explains why the best Top-1, Top-5, and MRR results for the filtered model are obtained for $D=40$~km.

\subsection{Comparative Pl@ntNet API Results} 
\label{results:plantnet}

To compare Floralens with the current state of the art in automatic plant identification, and given the restrictions in accessing the APIs for most of the platforms of interest (cf.\ discussion in Section ~\ref{sec:related}), we opted to use Pl@ntNet, for which we were generously granted free extended access. Recent work has identified Pl@ntNet as one of the best-performing platforms for this purpose~\cite{accuracyfree, mindyourapp} and, therefore, a good reference point for our evaluation.

The Pl@ntNet API~\cite{plantnet_api} is a web service that provides the same visual identification models used by Pl@ntNet apps~\cite{plantnet-model}. The API returns a set of ranked species for a given image for two models for worldwide flora: a so-called ``legacy'' model from 2022 (henceforth, \pnleg{}) generated using CNNs, and a more recent model announced in July 2023~\cite{plantnet_api_announcement}, generated using vision transformers (henceforth, \pnnew{}). Through the API, it is also possible to filter results from the \pnnew{} model so that only species occurring in a specific biogeographic region are included. One such region is Southwestern Europe, including Portugal, enabling the most straightforward comparison between Floralens and Pl@ntNet. These results are identified by \pnswe{}.

\begin{figure}[t!]
\centering
\includegraphics[width=\linewidth]{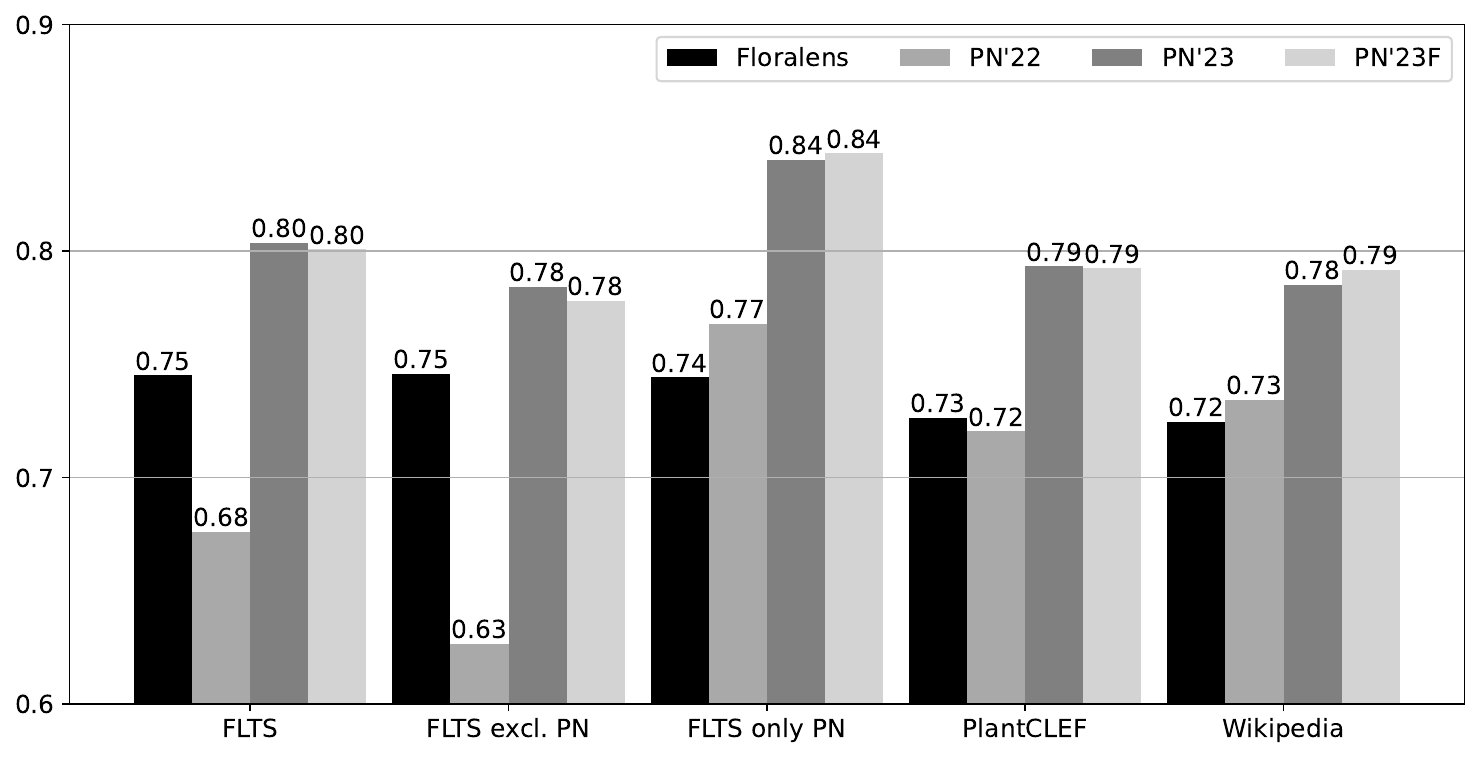}
\caption{Pl@ntNet API: \mrrs{} results as a function of dataset.\label{fig:plantnet_comparison_results}}
\end{figure}

Figure~\ref{fig:plantnet_comparison_results} shows a comparison of the \mrrs{} values obtained for Floralens, \pnleg{}, \pnnew{}, and \pnswe{} for all test datasets (FLTS, PlantCLEF, and Wikipedia). We also depict separate results gauging the impact of Pl@ntNet images: the \mrrs{} for the FLTS subset excluding Pl@ntNet images (cf.\ ``FLTS excl. PN'' series) and for the FLTS subset containing only Pl@ntNet images (cf.\ ``FLTS only PN'' series). Recall that images from Pl@ntNet are used to train our model, just as they are for the Pl@ntNet models themselves. So part of the Pl@ntNet images we use for testing may have been used to train the Pl@ntNet models.
Hence, the set of FLTS test images excluding Pl@ntNet may, in this sense, be a fairer comparison between Floralens and Pl@ntNet.

In the remainder of this section, we compare Floralens vs.\ Pl@ntNet results using differences in MRR values: \emph{a positive difference indicates that Floralens performs better than Pl@ntNet models, a negative difference the opposite}.

Regarding the overall test datasets, we can draw the following main conclusions:
\begin{enumerate} 
\item Floralens performs better than \pnleg{} by $+0.07$  ($0.75$ vs.\ $0.68$) for FLTS and has almost the same performance as \pnleg{} for PlantCLEF ($0.73$ vs.\ $0.72$), and Wikipedia ($0.72$ vs.\ $0.73$); 
\item Floralens performs worse than both \pnnew{} and \pnswe{}  by $-0.05$ for FLTS ($0.75$ vs.\ $0.80$), $-0.06$ for PlantCLEF ($0.73$ vs.\ $0.79$), and $-0.06$/$-0.07$ for Wikipedia ($0.72$ vs.\ $0.78$ and $0.72$ vs.\ $0.79$);
\item and the Southwestern Europe species filter associated with \pnswe{} has little impact on the results, compared to \pnnew{}.
\end{enumerate}

Turning our attention to the impact of Pl@ntNet images used in the FLTS, we can compare the “FLTS only PN” series (only Pl@ntNet images, $35$\% of the total FLTS) and the “FLTS excl. PN” series (no Pl@ntNet images, the remaining $65$\%) in Figure~\ref{fig:plantnet_comparison_results}. The main observations are:
\begin{enumerate}
\item the behavior of the Pl@ntNet models is strikingly different for both series, as the MRR of \pnleg{} improves from $0.63$ (non-Pl@ntNet images) to $0.77$ (Pl@ntNet images), 
and  \pnnew{}/\pnswe{} from $0.78$ to $0.84$;
\item in contrast, the corresponding scores for Floralens remain quite close across these two sets at $0.75$ and $0.74$, respectively.
\end{enumerate}
A positive bias  towards Pl@ntNet images by Pl@ntNet models thus seems evident. Hence, excluding Pl@ntNet images from FLTS may provide a more balanced comparison to Floralens,
and one that clearly favors Floralens. For non-Pl@ntNet images, Floralens performs much better than \pnleg{} by $+0.12$ ($0.75$ vs.\ $0.63$) and slightly worse than  \pnnew{}/\pnswe{} by $-0.03$ ($0.75$ vs.\ $0.78$).
This is in stark contrast with Pl@ntNet images, where Floralens performs worse by $-0.03$ ($0.74$ vs.\ $0.77$) than \pnleg{}, and by $-0.10$ ($0.74$ vs.\ $0.84$) than  \pnnew{}/\pnswe{}.

\begin{figure}[h!]
\centering
    \begin{subfigure}{0.9\linewidth}
        \centering
        \includegraphics[width=\textwidth]{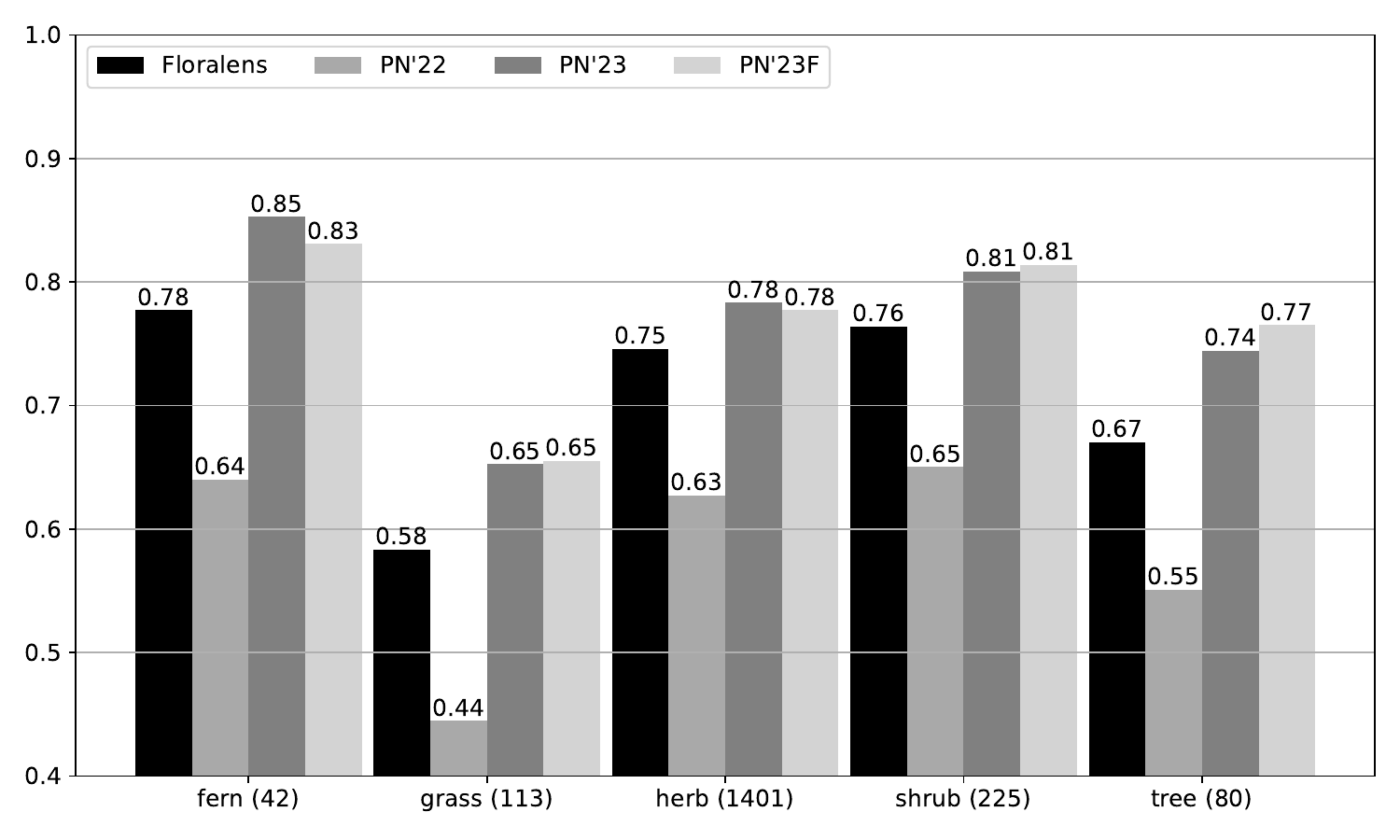}
        \subcaption{MRR results for Floralens and Pl@ntNet models for growth forms.}
        \label{fig:pn_gf_results}
    \end{subfigure} \\
    \begin{subfigure}{0.9\linewidth}
        \centering
        \includegraphics[width=\textwidth]{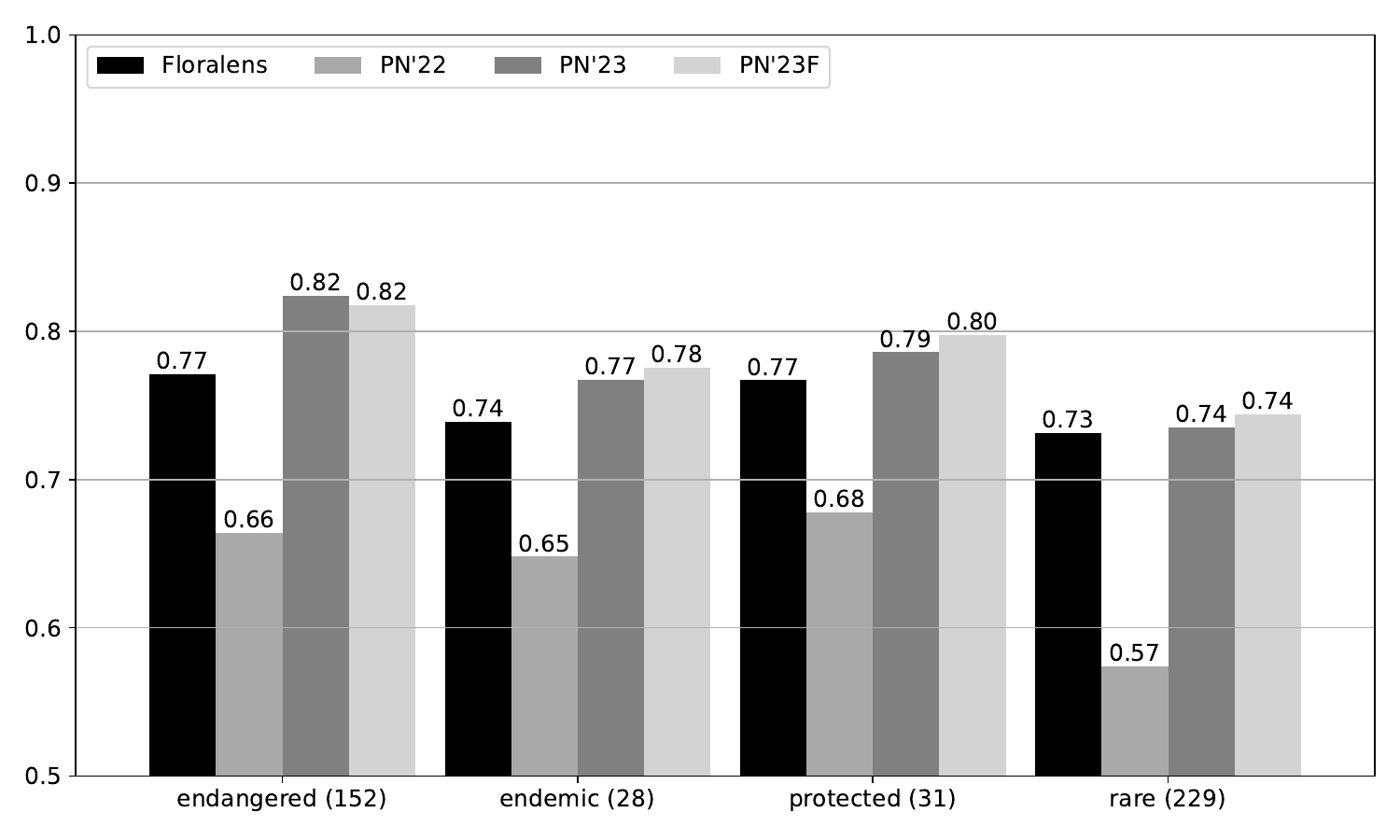}
         \subcaption{MRR results for Floralens and Pl@ntNet models for special categories.}
        \label{fig:pn_cs_results}
    \end{subfigure} 
    \caption{MRR results for Floralens and Pl@ntNet models.\label{fig:pn_gf_and_cs_results}}
\end{figure}

Finally, Figure~\ref{fig:pn_gf_and_cs_results} presents comparative MRR results again for FLTS, but grouped by (\ref{fig:pn_gf_results}) species growth type and (\ref{fig:pn_cs_results}) the special categories of species as analyzed previously for Floralens (cf.\ Section~\ref{results:baseline}). To achieve a more balanced comparison, in line with the above discussion, we exclude FLTS images obtained from Pl@ntNet. The results follow the overall trends discussed earlier. Floralens exceeds \pnleg{} by $+0.09$ for endemic species (Figure~\ref{fig:pn_cs_results}) and up to $+0.16$ for rare species. Conversely, it trails \pnnew{}/\pnswe{} by between $-0.01$ (rare species) and $-0.10$ (tree species). The most favorable comparisons occur for herb species ($0.75$ vs.\ $0.78$) and rare species ($0.73$ vs.\ $0.74$).

\section{Data and Software Artifacts}
\label{sec:artifacts}

In this section, we briefly describe the Biolens software platform, in which Floralens has been made publicly available, along with other artifacts provided to the community.

\subsection{Biolens Website}
The Floralens model has been integrated into the Biolens project website~\cite{biolens}. As illustrated in the screenshots of Figure~\ref{fig:app}, the functionality is quite simple: (a) users submit photos of interest, and (b) obtain identification suggestions with a textual indication of the model's confidence (high, medium, or low).
The app lists a suggestion for identification if a minimum threshold score of 15\% is reached, and at most five suggestions are listed. The website is hosted on a small virtual machine with just 2~CPU cores and 8~GB of RAM. With this configuration, invoking the model takes an average of 900~ms per image. This value was directly calculated from the approximately 29,000 requests to the server between April 2021 and April 2025. As for the model itself, it takes only 8~MB of disk storage. The server configuration is lightweight, as we utilize the TF Lite variant of the Floralens model that runs on the server side, similarly to models for other taxa hosted on the same site. The user's browser displays the server's results; it does not host the model.

\begin{figure}[t!]
\centering
\begin{subfigure}{0.475\linewidth}
       \vspace*{0.5cm}
        \centering
        \includegraphics[width=\textwidth]{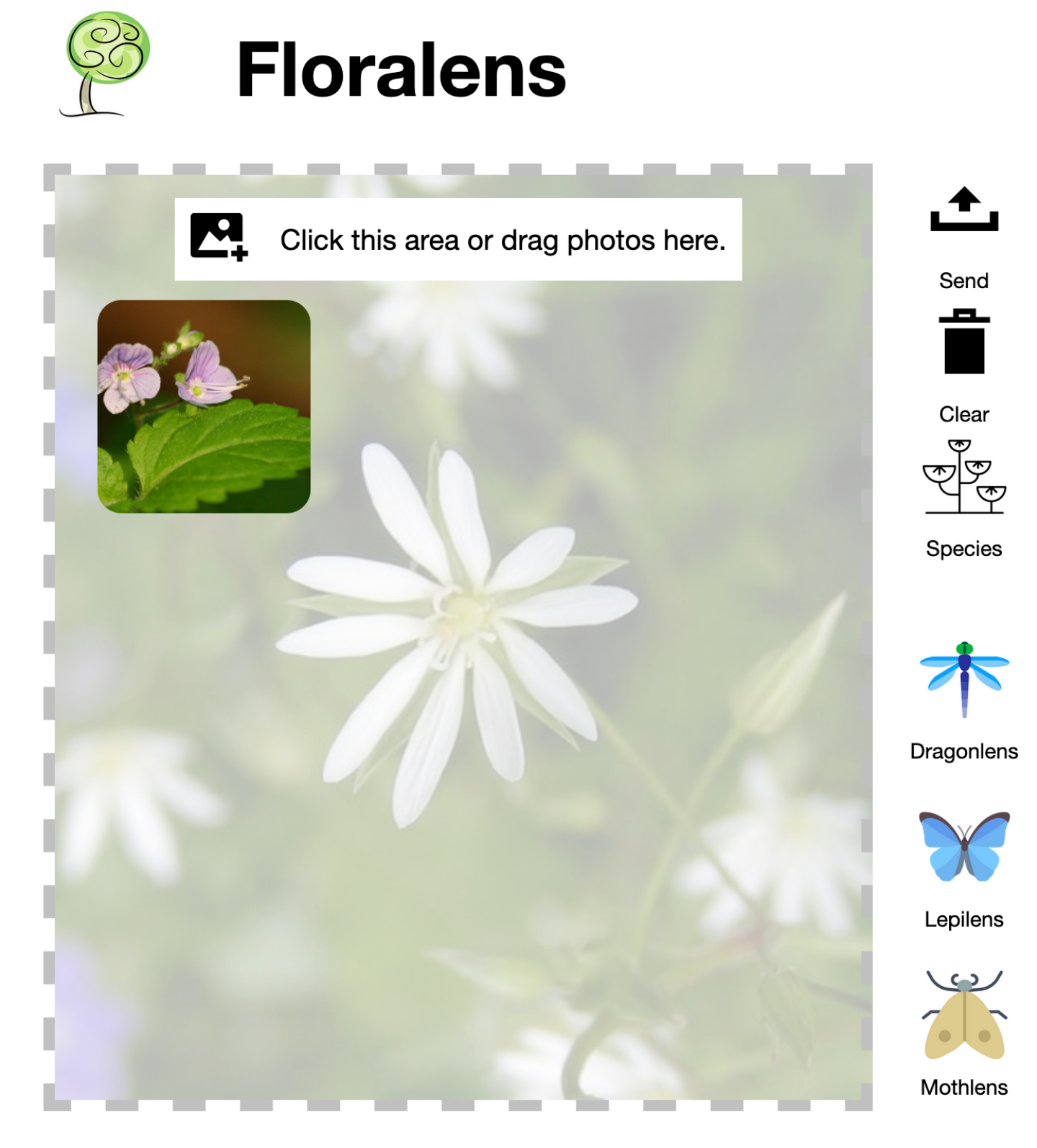}
        \subcaption{Image uploading by the user.}
        \label{fig:app:upload}
    \end{subfigure}
    \hfill
    \begin{subfigure}{0.475\linewidth}
        \centering
       \includegraphics[width=\textwidth]{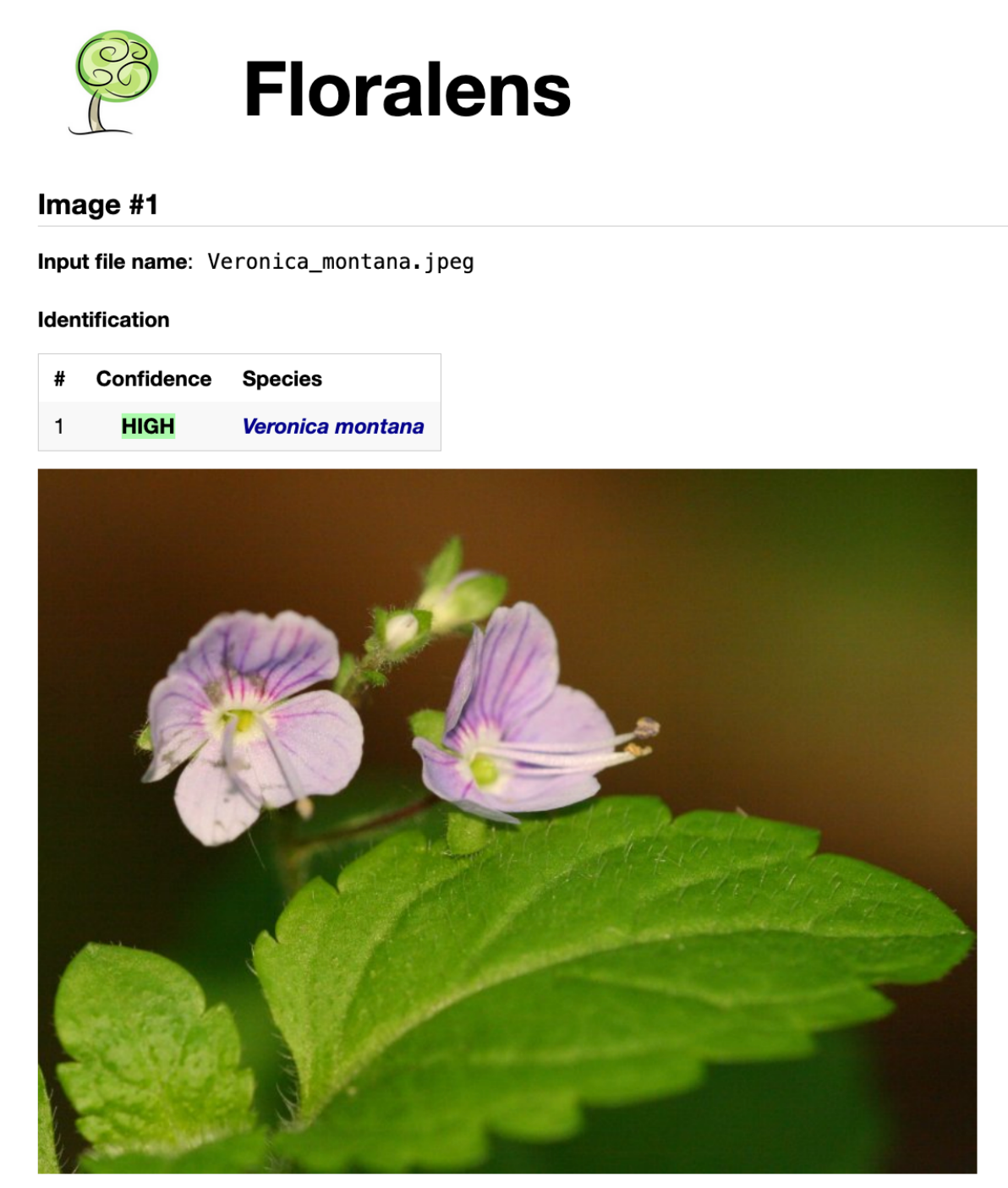}
        \subcaption{Suggested automatic identification.}
        \label{fig:app:results}
    \end{subfigure}  
\caption{Biolens -- web application screenshots.}\label{fig:app}
\end{figure}

\subsection{Biolens app}

We recently developed a prototype version of a mobile application that runs on Android and iOS devices. \anon{The Android version is available for download at the Biolens website.}{} A few screenshots of the application are shown in Figure~\ref{fig:mobile-app}. The functionality is similar to that of the Biolens website, but adapted for mobile use: users can take photos of specimens on the fly and obtain instant identification suggestions without an internet connection. Various models are bundled within the app and, thus, are evaluated \emph{in situ} on the mobile device. The identification information, date, current location, and optional user annotations are recorded in association with each photo. According to latency benchmarks provided by Google AutoML for  modern mobile devices like Google Pixel 2, Samsung Galaxy S7, or iPhone X, invoking the model takes less than 100~ms. This is consistent with our experience and is much faster than the web server discussed above, since the server configuration is considerably more lightweight than that of modern mobile devices.

\begin{figure}[t!]
\centering
\begin{subfigure}{0.3\linewidth}
        \centering
        \includegraphics[width=\textwidth]{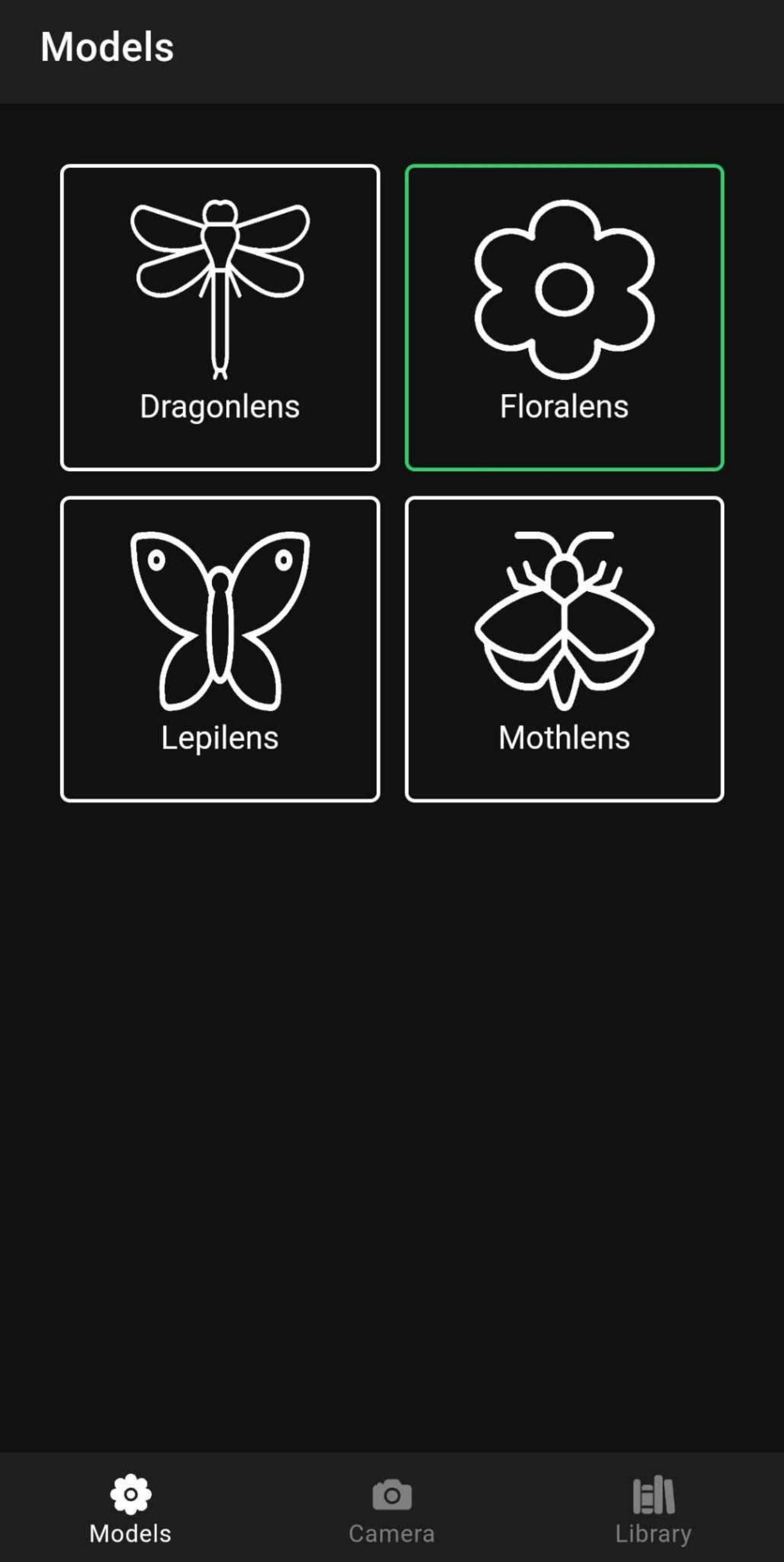}
        \subcaption{Model selection.}
        \label{fig:mapp:model-selection}
    \end{subfigure}
    \hfill
    \begin{subfigure}{0.3\linewidth}
        \centering
       \includegraphics[width=\textwidth]{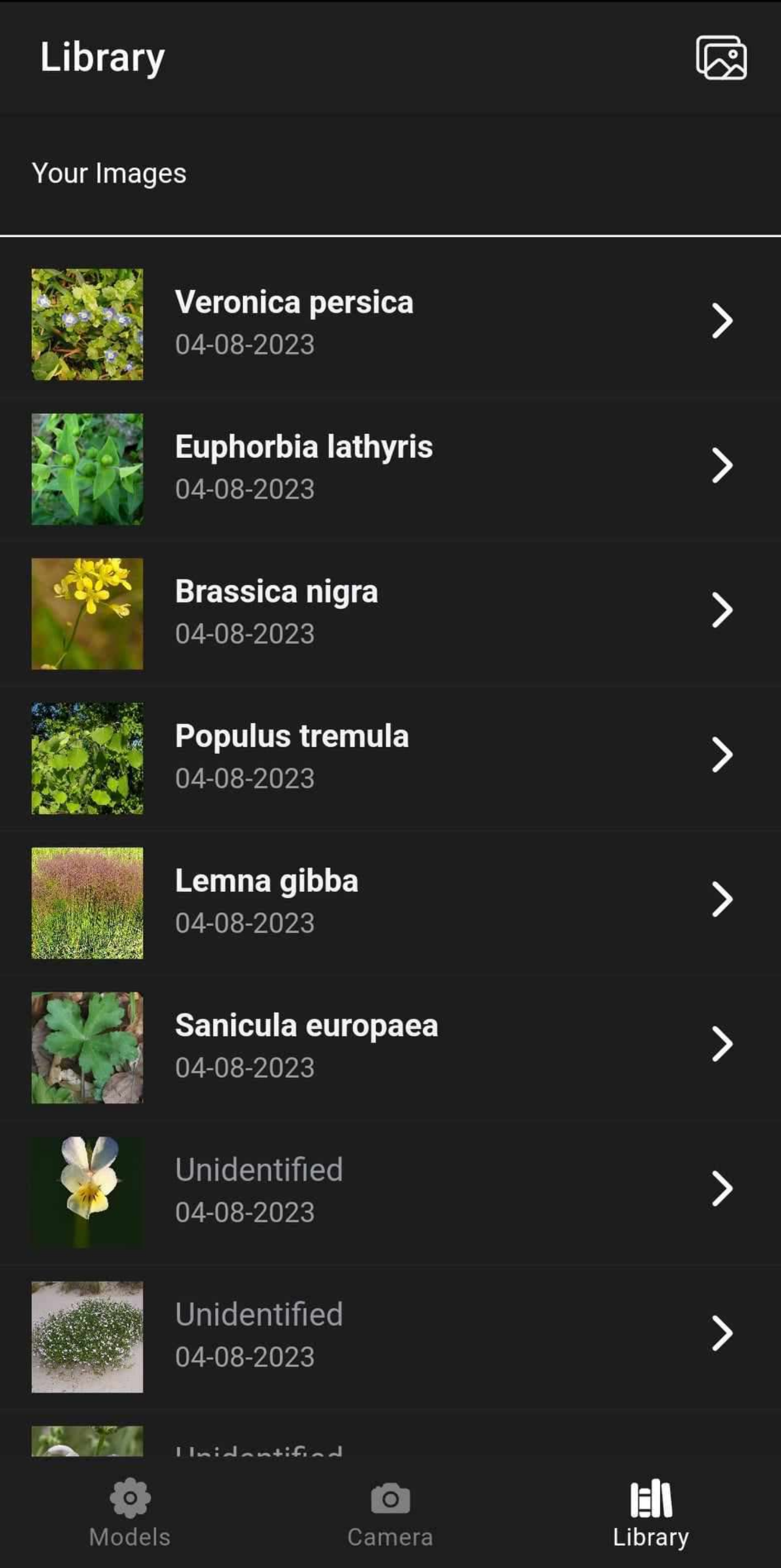}
        \subcaption{Image list.}
        \label{fig:mapp:library}
    \end{subfigure}  
     \hfill
    \begin{subfigure}{0.3\linewidth}
        \centering
       \includegraphics[width=\textwidth]{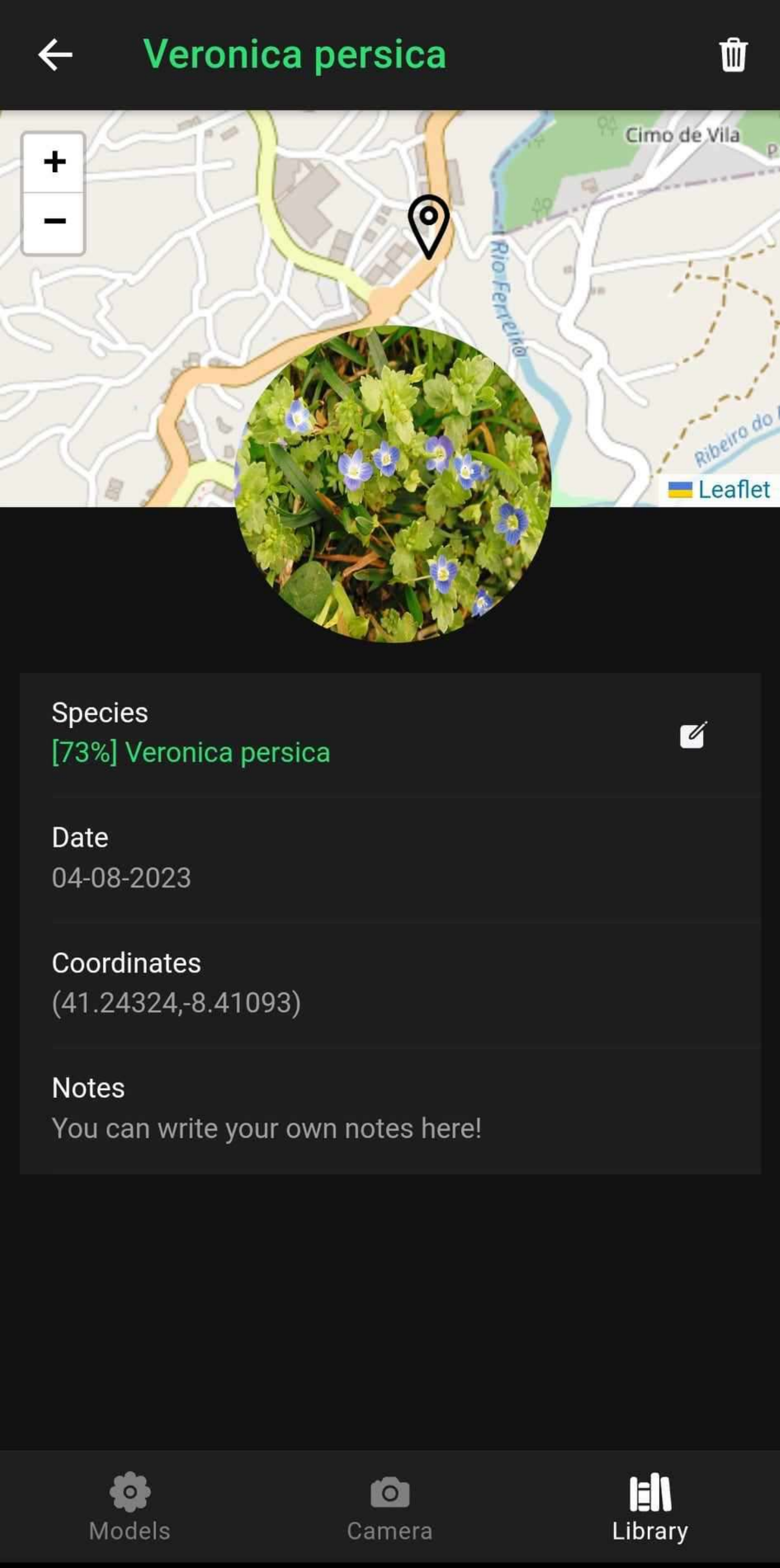}
        \subcaption{Details for an image.}
        \label{fig:mapp:details}
    \end{subfigure} 

\caption{Biolens -- mobile application screenshots.}\label{fig:mobile-app}
\end{figure}

\subsection{Dataset and Results}
The dataset and the Python notebooks with all the code used for the analysis presented in Section~\ref{sec:results} are 
available publicly from Zenodo~\cite{dataset}. The dataset contains: (a) a mapping between image identifiers and their ground-truth species labels; (b) the image URLs from which they were retrieved; (c) URLs for a repository we maintain, where all images are mirrored; (d) GBIF identifiers when applicable (all images except those obtained from FloraOn). Ground truth and URLs are also available for the PlantCLEF and Wikipedia datasets. The Top-5 predictions and corresponding confidence levels for the Floralens model and all three Pl@ntNet variants are included for all datasets.

\section{Discussion}
\label{sec:conclusions}

In this paper, we describe the construction of a dataset of Portuguese native flora and the derivation of a deep learning model for the automated identification of species within it. The species universe was taken from the FloraOn dataset, provided by the Sociedade Portuguesa de Botânica and compiled exclusively by specialists. The dataset was constructed by sampling high-quality data from several research-grade datasets available through GBIF. In addition to FloraOn, the data were sourced from iNaturalist, Pl@ntNet, and Observation.org. The Floralens dataset is available publicly on Zenodo~\cite{dataset}. The Floralens model was derived from this dataset using GAMLV, which enables users to train models using off-the-shelf CNNs. The model is available online at the Biolens Project website, and can also be used in the Biolens mobile application, for example, when working offline during fieldwork.

The baseline Floralens model has good predictive power, with an average precision of $0.72$. The precision and recall values for a reference confidence level of $0.5$ are $0.85$ and $0.53$, respectively. It also exhibits  relatively homogeneous predictive performance across all the dataset's sources. We measured values of $0.72$ for Top-1, $0.88$ for Top-5, and $0.78$ for MRR. 
An interesting result concerns the model's accuracy in identifying endangered, rare, protected, and endemic species. Some research (e.g.,~\cite{mindyourapp}) suggests that current platforms struggle with these classes. The results for Floralens in these categories show that the model retained its nominal predictive performance.
Using multiple images of the same specimen significantly improved the model's accuracy for all metrics: $0.80$ for Top-1, $0.94$ for Top-5, and $0.86$ for MRR. This agrees with the findings reported by other projects (e.g.,~\cite{rzanny2019,plantnet-model}).
Introducing filters based on species' geographical distribution also improved the model's accuracy, though not substantially. The results attest to the accuracy and completeness of the FloraOn dataset's geographic information. 
Finally, the accuracy of the Floralens model is higher than Pl@ntNet's legacy model, which is also based on CNNs. Pl@ntNet's latest model, which employs vision transformers, is more accurate, although the gap is small. In other words, for the same model technology, Floralens outperforms Pl@ntNet for the Portuguese flora.

As for future work, we aim to improve the model's accuracy by increasing the number of observations in the dataset that feature multiple images of the same specimen, ideally tagged by plant part. This will involve a concerted effort  with users of the Floralens app and botanical specialists, combined with extracting additional multi-image observations from yet-unexplored public datasets (e.g., Flora Incognita). Including images from Morocco and Algeria will enrich the dataset with additional species from the broader Mediterranean flora. We also intend to address limitations arising from inconsistent use of taxonomic names and synonyms.
Another direction worth pursuing is using vision transformers to train our models. As noted in Section~\ref{sec:related}, this is a relatively recent technology that may yield higher accuracy than the use of CNNs. 
We also plan to continue preliminary work on using image similarity models~\cite{filgueiras2022}. These may provide an alternative way of confirming identifications when traditional computer vision models yield low-confidence results. Hybrid classification models that combine both approaches are a possibility worth exploring.
Finally, further work is required on the Biolens mobile app to improve its usability, optimize resource usage, and integrate it with existing citizen science platforms.

}
\paragraph{Acknowledgements.} \nrev{The authors thank Miguel Porto and Henrique Alves for their help with the FloraOn dataset and numerous insights, and Hugo Gresse, Pierre Bonnet, and Mathias Chouet for kindly giving us extended access to the Pl@ntNet API.
This work was partially funded by projects SafeCities and Augmanity (POCI-01-0247-FEDER-041435 and -046103, through COMPETE 2020 and Portugal 2020),  by project UIDB/50014/2020 (Fundação para a Ciência e Tecnologia), and a cloud resource credits through the Google Cloud Research Credits program.} 

\anon{}{
\bibliographystyle{unsrt}
\bibliography{refs}
}

\end{document}